%% file: neurips_2026.tex
\documentclass{article}

\input{preamble}

\title{Skill-Aligned Annotation for Reliable \\ Evaluation in Text-to-Image Generation}

\author{%
  Abdelrahman Eldesokey \quad Merey Ramazanova \quad Ahmad Sait \quad Ansar Khangeldin \\
  \textbf{Karen Sanchez \quad Tong Zhang \quad Bernard Ghanem} \\ \\  
  King Abdullah University of Science and Technology (KAUST) \\
  Thuwal, Saudi Arabia \\
  \texttt{\{firstname.lastname\}@kaust.edu.sa}
}

\begin{document}

\maketitle

    

\input{sec/0_abstract}

\input{sec/1_intro}

\input{sec/2_related}

\input{sec/3_skill_tax}
\input{sec/4_annot_stategy}
\input{sec/5_full_eval}

\input{sec/6_automated}

\input{sec/7_conclusion}

\bibliographystyle{abbrv} 
\bibliography{bibliography}


\newpage

\appendix
\input{sec/appendix}



\end{document}

%% file: preamble.tex
    \PassOptionsToPackage{numbers, compress}{natbib}
 \usepackage[preprint]{neurips_2026}

\usepackage[utf8]{inputenc} 
\usepackage[T1]{fontenc}    
\usepackage{hyperref}       
\usepackage{url}            
\usepackage{booktabs}       
\usepackage{amsfonts}       
\usepackage{nicefrac}       
\usepackage{microtype}      
\usepackage{xcolor}         
\usepackage{booktabs}
\usepackage{multirow}
\usepackage{tabularx}
\usepackage{array}
\usepackage{fontawesome5}

\usepackage{tcolorbox}
\usepackage{listings}
\usepackage{xcolor}
\usepackage{xspace}
\usepackage{amsmath}
\usepackage{graphicx}
\usepackage{subcaption}
\usepackage{longtable}
\usepackage{caption}

\definecolor{lightgraybg}{RGB}{245,245,245}
\definecolor{commentgreen}{RGB}{0,120,0}

\lstdefinestyle{jsonstyle}{
    backgroundcolor=\color{lightgraybg},
    basicstyle=\ttfamily\scriptsize,
    frame=single,
    framerule=0pt,
    rulecolor=\color{lightgraybg},
    xleftmargin=2pt,
    xrightmargin=2pt,
    aboveskip=4pt,
    belowskip=4pt,
    breaklines=true,
    showstringspaces=false,
    commentstyle=\color{commentgreen},
    morecomment=[l]{\#}   
}

\usepackage[capitalize]{cleveref}
\crefname{section}{Sec.}{Secs.}
\Crefname{section}{Section}{Sections}
\Crefname{table}{Table}{Tables}
\crefname{table}{Tab.}{Tabs.}
\crefname{appendix}{appendix}{appendices}
\Crefname{appendix}{Appendix}{Appendices}

\crefname{subappendix}{appendix}{appendices}
\Crefname{subappendix}{Appendix}{Appendices}

\AddToHook{cmd/appendix/before}{%
  \crefalias{section}{appendix}%
  \crefalias{subsection}{subappendix}%
}

\def\eg{\textit{e.g.}\@\xspace}

\newcommand{\myparagraph}[1]{{ \noindent \textbf{#1}}}
\newcommand{\new}[1]{{\color{black}#1}}

\newif\ifshowcomments
\showcommentstrue   

%% file: sec/0_abstract.tex
\begin{abstract}


Text-to-image (T2I) generation has advanced rapidly, making reliable evaluation critical as performance differences between models narrow.
Existing evaluation practices typically apply uniform annotation mechanisms, such as Likert-scale or binary question answering (BQA), across heterogeneous evaluation skills, despite fundamental differences in their nature.
In this work, we revisit T2I evaluation through the lens of \emph{skill-aligned} annotation, where annotation strategies reflect the underlying characteristics of each evaluation skill.
We systematically compare skill-aligned annotation against uniform baselines and show that it produces more consistent evaluation signals, with higher inter-annotator agreement and improved stability across models.
Finally, we present an automated pipeline that instantiates the proposed evaluation protocol, enabling scalable and fine-grained evaluation with spatially grounded feedback.
Our work highlights that improving the foundations of image evaluation can increase reliability and efficiency without simply scaling annotation effort.
We hope this motivates further research on refining evaluation protocols as a central component of reliable model assessment.
\end{abstract}

%% file: sec/1_intro.tex
\section{Introduction}

Text-to-Image (T2I) generation has recently undergone rapid progress with the emergence of diffusion models \cite{ho2020denoising,song2021maximum,imagen2022,podell2024sdxl,chen2024pixartalpha,esser2024scaling}. 
As performance differences between models continue to narrow, \emph{reliable evaluation} becomes increasingly critical for guiding meaningful progress and identifying failure modes. 
A large body of work has therefore focused on evaluating T2I systems across multiple axes, including semantic fidelity, compositional correctness, and visual quality \cite{kirstain2023pick,xu2023imagereward,wu2023human,lee2023holistic,lin2024evaluating,hu2023tifa,chao2024dsg,ghosh2023geneval,wiles2025revisiting,hayes2025finegrain,saxon2024who}.
In practice, the gold standard for T2I evaluation relies on human feedback, where human annotators assess different evaluation skills (\eg spatial arrangement, text rendering, visual quality) using \emph{uniform annotation mechanisms}, such as Likert-scale ratings, ranking, or binary question answering (BQA). 
However, these mechanisms are typically applied uniformly across heterogeneous skills, despite fundamental differences in their structure. 
The recent work Gecko \cite{wiles2025revisiting}) showed that annotation design can significantly affect both inter-annotator agreement and evaluation outcomes, raising a central question: ``\emph{Are standard annotation mechanisms uniformly effective in capturing heterogeneous skills in T2I generation?}''

\begin{figure}
    \centering
    \includegraphics[width=\linewidth]{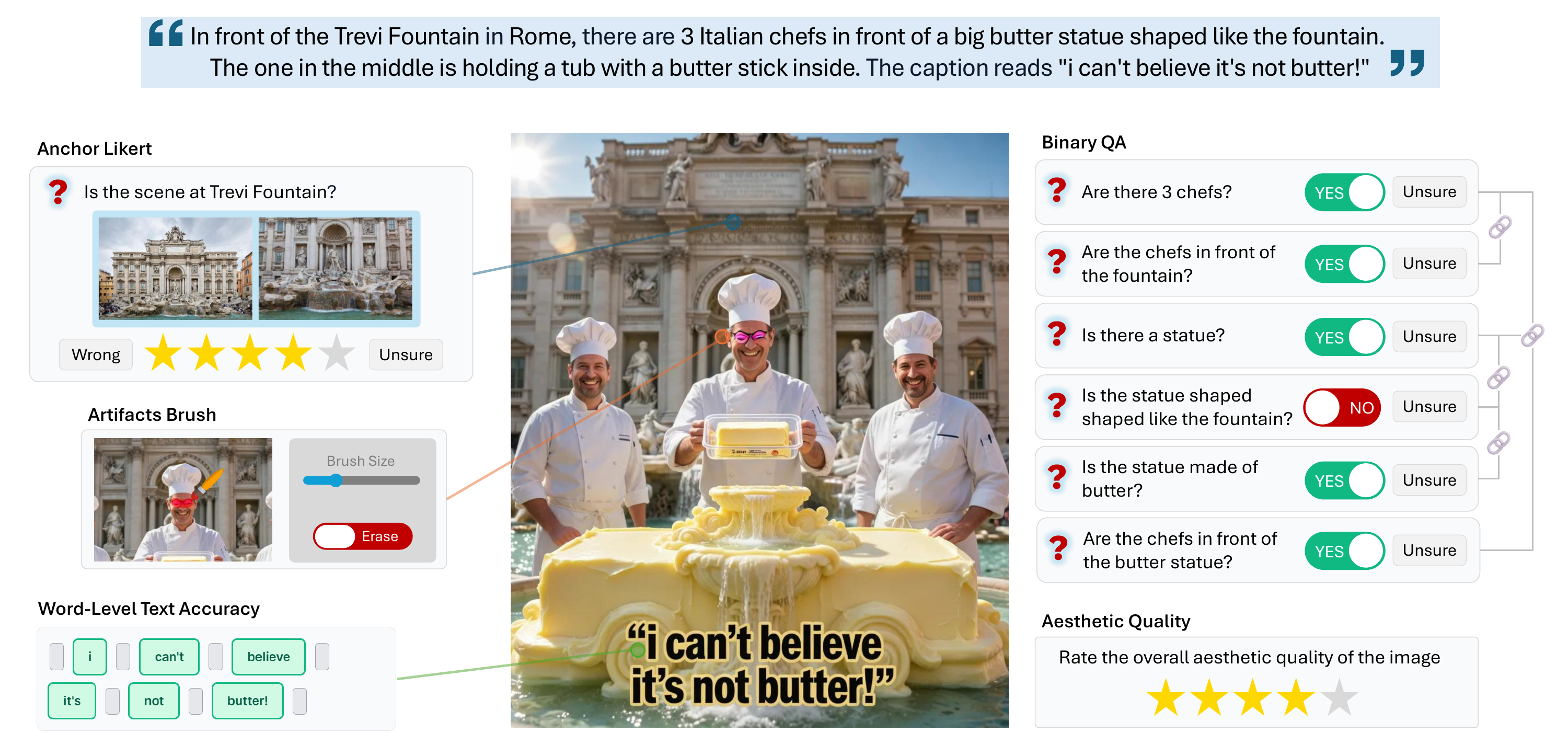}
    \caption{Overview of the proposed Text-to-Image evaluation protocol using \emph{skill-aligned} annotation, yielding more consistent and stable evaluation signals than uniform baselines.}
    \label{fig:teaser}
\end{figure}

Current evaluation pipelines implicitly assume that a single annotation strategy can capture diverse evaluation skills. This assumption is often invalid in practice. 
For example, assessing landmark or style recognition via BQA presumes annotators possess sufficient prior knowledge, which is rarely verified. 
Conversely, compressing spatially localized artifacts into a single Likert score discards fine-grained evidence and can lead to disproportionate penalties for minor defects. 
These mismatches suggest that \emph{annotation design is a critical but underexplored component of evaluation methodology}.

In this work, we revisit T2I evaluation from the perspective of \emph{skill-aligned} annotations, where annotation mechanisms are chosen to match to the nature of each evaluation skill. 
To achieve this, we first define a taxonomy of evaluation skills spanning semantic and aesthetic dimensions, and use it to systematically tag image generation text prompts with different skills. 
Building on this taxonomy, we design annotation strategies tailored to different skill types.

Crucially, we treat annotation design as a \emph{methodological variable} and conduct controlled comparisons between skill-aligned and uniform annotation protocols. 
We show that skill-aligned annotation produces more consistent stable outcomes that converge with a few number of annotators, while achieving higher inter-annotator agreement. 
Finally, to support large-scale evaluation, we instantiate the proposed protocol in an automated pipeline that enables scalable, fine-grained, and spatially grounded assessment, achieving strong alignment with human judgments across most skills.


We hope this work encourages a shift toward more principled evaluation design in T2I generation, emphasizing the role of annotation protocols in determining evaluation reliability \footnote{The source code is available at \href{https://github.com/abdo-eldesokey/skill-aligned-eval}{https://github.com/abdo-eldesokey/skill-aligned-eval}}. 


%% file: sec/2_related.tex
\vspace{-7pt}
\section{Related Work}


\subsection{Evaluation Axes of T2I Generation}

Evaluating T2I models requires defining axes that capture distinct aspects of image quality. 
Early and widely adopted approaches \cite{kirstain2023pick,wu2023human,wu2023humanv2,ma2025hpsv3} formulate evaluation as a global human preference task, where annotators rank images according to overall appeal. 
While such signals capture aggregate user preference, they do not disentangle the underlying factors that contribute to image quality, limiting their usefulness for diagnosing model behavior or enforcing task-specific requirements.
To address this limitation, a large body of work focuses on more targeted evaluation axes, most notably \emph{semantic fidelity} to the input prompt. 
These approaches assess whether generated images satisfy explicit textual constraints, such as object presence, counts, attributes, and spatial relationships \cite{chao2024dsg,lin2024evaluating,hu2023tifa,ghosh2023geneval,wiles2025revisiting,li2024genaibench,huang2025t2i,lu2023llmscore,ku2024viescore,zhang2025strict}. 
Such formulations improve diagnostic granularity but often abstract away perceptual qualities of the image. 
Complementary work therefore considers perceptual axes, including realism, aesthetics, and visual artifacts, either independently or jointly with semantic fidelity \cite{zhang2024learning,otani2023toward,hua2025mmig,xu2023imagereward,xu2024visionreward,lee2023holistic}.

As T2I models continue to improve and performance gaps narrow, the primary challenge is no longer the introduction of additional evaluation axes, but the design of evaluation protocols that measure these axes in a consistent and reliable manner. 
In contrast to prior work that focuses on \emph{what} to evaluate, we focus on \emph{how} to evaluate each axis, and propose a skill-aligned evaluation methodology that adapts annotation protocols to the structure of each evaluation skill.


\subsection{Annotation Strategies in T2I Evaluation}

Human evaluation remains the gold standard for assessing T2I generation, and most prior work relies on large-scale user studies to collect evaluation signals. 
A common approach is \emph{comparative} evaluation, where annotators choose between images generated by different models \cite{kirstain2023pick,wu2023human,wu2023humanv2,ma2025hpsv3}. 
Such comparisons reduce absolute calibration requirements and often yield consistent judgments. 
However, comparative evaluation is inherently \emph{relative}: it depends on the choice of baselines and does not provide direct insight into which specific aspects of an image drive a preference \cite{otani2023toward}. 
Moreover, as the number of models increases, pairwise comparisons scale quadratically and become prohibitively expensive, while offering limited diagnostic value for identifying failure modes.



For these reasons, we focus on \emph{absolute} evaluation, where annotators assess individual images against predefined criteria using mechanisms such as Likert-scale ratings or binary question answering (BQA) \cite{chao2024dsg,hu2023tifa,ghosh2023geneval,xu2023imagereward,wiles2025revisiting,saxon2024who}. 
Unlike pairwise comparisons, this paradigm supports fine-grained, skill-level analysis and direct attribution of errors, while avoiding dependence on explicit baselines. 
However, its effectiveness depends strongly on annotation mechanism design, since uniform protocols may not measure heterogeneous skills with equal reliability.


\subsection{Automated Evaluation of T2I Models}

To reduce the cost and latency of human evaluation, prior work seeks to automate T2I assessment using models trained on signals derived from user studies. 
A common approach is to fine-tune CLIP-based models or related architectures as preference or scoring functions \cite{kirstain2023pick,xu2023imagereward,wu2023humanv2,zhang2024learning}, or to employ learned models to answer task-specific VQA-style queries that probe semantic correctness \cite{chao2024dsg,hu2023tifa,li2024genaibench,lu2023llmscore}. 
While these methods improve scalability, their behavior is inherently tied to the annotation protocols and biases present in the underlying human data.
More recent work leverages large Vision--Language Models (VLMs) as general-purpose evaluators \cite{xu2024visionreward,ma2025hpsv3,hua2025mmig,hayes2025finegrain,tu2025automatic}. 
These models can flexibly score images, rank candidates, or answer diverse evaluation queries without task-specific retraining. 
However, their outputs often lack interpretability, conflate multiple evaluation axes, and remain sensitive to prompt formulation, limiting control over what is being measured.

Importantly, prior work often assumes that automated evaluators can be applied uniformly across evaluation axes. 
MMIG-Bench \cite{hua2025mmig} partially challenges this assumption by using different models for low- and high-level properties, suggesting that evaluation mechanisms should be aligned with the nature of each axis. 
Following this direction, we instantiate our protocol in an automated pipeline that assigns different mechanisms or models to different evaluation skills.


%% file: sec/3_skill_tax.tex
\section{Evaluation Setup for Text-to-Image Generation}

We study the interaction between \emph{annotation strategies} and \emph{evaluation skills} in Text-to-Image (T2I) evaluation, with the goal of understanding how annotation design affects evaluation outcomes. 
We hypothesize that applying a uniform annotation strategy across heterogeneous skills introduces avoidable subjectivity and obscures skill-specific failure modes. 
To investigate this, we define a structured taxonomy that decomposes T2I performance into semantic and aesthetic skills, and associate each skill with candidate annotation strategies. 
We then conduct controlled, per-skill comparisons of annotation protocols and evaluate their reliability using inter-annotator agreement (\eg, Krippendorff’s $\alpha$), disagreement rates, and convergence behavior. 
Based on these findings, we construct an evaluation suite that adopts skill-aligned annotation strategies, and instantiate it in an automated pipeline for scalable, standardized evaluation.


\subsection{Skill Taxonomy for T2I Evaluation}
\label{sec:taxonomy}

\input{sec/skill_tax_table}

Several prior works propose taxonomies for evaluating text-to-image (T2I) generation~\cite{chao2024dsg,hu2023tifa,wiles2025revisiting,lin2024evaluating}. 
Building on these efforts, we construct a taxonomy that captures the key skills required for high-quality image generation, spanning both \emph{semantic fidelity} and \emph{aesthetic quality}. 
\Cref{tab:evaluation_skills} summarizes the taxonomy, and per-skill examples are provided in \Cref{sec:skill_details}.
For evaluation prompts, we adopt the Gecko prompt set~\cite{wiles2025revisiting} due to its broad coverage across existing benchmarks. 
In contrast to Gecko, which assigns a single skill--subskill pair per prompt, we annotate each prompt with all applicable skill and sub-skill combinations. 
This design reduces the number of prompts required to cover the taxonomy and enables analysis of model performance under combinations of skills.


\subsection{Skill Tagging}

Prior work explored tagging prompts using large language models (LLMs)~\cite{hu2023tifa,chao2024dsg,li2024genaibench}, followed by a second LLM that generates (BQA) probes to validate skill fulfillment.
We adopt a similar approach but jointly perform tagging and question generation within a single LLM prompt. 
We found that recent LLMs (\eg GPT-5) can perform both tasks simultaneously given a well-designed system prompt.
For each input prompt, the LLM is provided with the full skills taxonomy and asked to:
(1) identify all relevant skill--subskill pairs, and  
(2) generate validation questions.

Following DSG~\cite{chao2024dsg}, we structure the questions as dependency nodes. 
For example, in the prompt ``A red car'', the color question depends on the presence of the car.
The LLM is instructed to produce structured output in the following format:

\begin{lstlisting}[style=jsonstyle]
uid: str                # Unique ID per skill-subskill
skill: str              # Representative skill
subskill: str           # Representative subskill (empty if none)
phrase: str             # Phrase in the prompt representing the skill
question: str           # Binary QA validating the skill
node_type: str          # "presence", "property", or "relation"
depends_on: List[str]   # List of parent UIDs
\end{lstlisting}

An overview of the tagging pipeline is shown in \Cref{fig:tagging}.
More details are provided in \Cref{sec:tagging_app}.


%% file: sec/skill_tax_table.tex
\begin{table*}[!t]
\centering
\footnotesize
\setlength{\tabcolsep}{4pt}          
\renewcommand{\arraystretch}{1.2}    

\begin{tabularx}{\textwidth}{
>{\raggedright\arraybackslash}m{3.2cm}
>{\raggedright\arraybackslash}X
>{\raggedright\arraybackslash}m{2.5cm}
}
\toprule
\textbf{Evaluation Skill} & \textbf{Sub-Skills} & \textbf{Annotation} \\
\midrule

\faCube\ Entities
& Singular / Count / Uncountable
& \multirow{12}{=}{BQA} \\

\faSlidersH\ Attributes
& Color / Texture / Material / Shape / Scale
& \\

\faRunning\ Action
& Standard / Unusual / Pose
& \\

\faProjectDiagram\ Arrangement
& --
& \\

\faBalanceScale\ Comparison
& Scale / Tone / Distance / Count / Other
& \\

\faLightbulb\ Lighting
& --
& \\

\faCloudSun\ Weather
& --
& \\

\faEye\ View
& --
& \\

\faCamera\ Camera
& --
& \\

\faSmile\ Mood / Feeling
& --
& \\

\faLanguage\ Language Complexity
& Negation / Color Stroop
& \\

\faClock\ Time
& Time of Day / Season / Year / Era
& \\

\cmidrule(lr){1-3}

\faMountain\ Environment / Scene
& Landmark / General
&  \multirow{3}{=}{Anchor BQA/Likert} \\

\faPaintBrush \ Style 
& Artistic Style / Visual Medium
&  \\

\faIdBadge\ Named Entities
& Character / Vehicle / Product / Artwork
&  \\

\cmidrule(lr){1-3}

\faFont\ Text Rendering
& Accuracy / Style / Numerical / Position
&  Likert/Word-Level \\

\cmidrule(lr){1-3}

\faExclamationTriangle\ Artifacts
& --
&  Likert/Brush \\

\faStar\ Aesthetic Quality
& --
& Likert \\

\bottomrule
\end{tabularx}
\vspace{5pt}
\caption{Skill taxonomy with proposed annotation strategies for each skill. We explain Brush annotation in \Cref{sec:visual_artifacts}, Word-Level in \Cref{sec:text_rendering}, and Anchor BQA/Likert in \Cref{sec:anchor_based}.}
\label{tab:evaluation_skills}
\end{table*}

%% file: sec/4_annot_stategy.tex
\section{Skill-Aligned Annotation}

Given the skill taxonomy introduced in \Cref{sec:taxonomy}, we study how annotation protocols interact with different T2I skills. 
Our goal is to test whether protocols aligned with each skill provide more reliable evaluation signals than uniform annotation strategies. 
We quantify reliability using inter-annotator agreement, disagreement rates, and convergence as a function of annotator count.

As summarized in \Cref{tab:evaluation_skills}, \emph{factual skills} such as object presence, counting, spatial arrangement, and temporal attributes are evaluated with binary question answering (BQA) as it can be adequately captured with Yes/No answer.. 
For skills with less discrete evidence, such as visual artifacts, text rendering, landmarks, and artistic style, we compare multiple candidate annotation strategies. 
We recruit six professional annotators with postgraduate degrees in CV/ML, balanced by gender and drawn from diverse cultural backgrounds, to evaluate skill-targeted prompts under controlled conditions. 
The same annotators participate in all conditions, enabling within-subject comparisons and reducing confounds from annotator variation. 
For model diversity, we generate images using three open-source T2I models: FLUX.1-dev~\cite{flux2024}, Z-Image-Turbo~\cite{cai2025z}, and FLUX.2-dev~\cite{flux-2-2025}.


\subsection{Visual Artifacts}
\label{sec:visual_artifacts}
Visual artifacts are spatially localized but are often evaluated using global Likert ratings for realism or artifact severity~\cite{xu2023imagereward,zhang2024learning}. 
To evaluate the reliability of this approach, we sample 30 prompts that commonly induce artifacts, including humans, vehicles, animals, actions, and text rendering.
For each prompt, we generate images with the three models described above and ask annotators to rate each image on a 1--5 Likert scale reflecting perceived realism and absence of artifacts.

\new{\Cref{fig:visual_artifacts_b} shows that Likert scoring yields low agreement, with Krippendorff's $\alpha = 0.39$ and $95\%$ bootstrap confidence intervals (CI) of $[0.27,\,0.50]$.}
To further characterize disagreement, we define the \textbf{Extreme Disagreement Rate (EDR)} as the proportion of images for which the maximum annotator score difference exceeds a threshold $\tau$. 
Let $D_i$ denote the difference between the maximum and minimum score assigned to image $i$. 
We compute
\begin{equation}
    \text{EDR} = \frac{1}{|V|} \sum_{i \in V} \mathbb{I}(D_i \ge \tau),
\end{equation}
where $V$ is the set of images annotated by at least two annotators and $\mathbb{I}(\cdot)$ is the indicator function. 
We set $\tau$ to $40\%$ of the scoring range, corresponding to two points on a 1--5 Likert scale. 
Under Likert scoring, the EDR reaches $62.2\%$, indicating substantial disagreement among annotators.


\myparagraph{Brush-Based Artifact Annotation}
Because artifacts are spatially localized, we compare Likert scoring with region-based brush annotation, which is commonly used in artifact localization~\cite{eldesokey2025mindtheglitch,kang2025legion,zhang2023perceptual}. 
Instead of assigning a global score, annotators mark image regions they perceive as artifacts. 
We convert each mask into a scalar artifact ratio by dividing the number of marked pixels by the total number of image pixels.

\Cref{fig:visual_artifacts_b} shows that brush-based annotation substantially improves reliability, increasing agreement to $\alpha = 0.82$ with CI $[0.56,\,0.89]$. 
The confidence intervals are disjoint from those of Likert scoring, indicating that the gain is statistically meaningful at this sample size. 
The EDR also drops to $0\%$, showing that annotators make more consistent judgments about artifact presence and extent. 
These results suggest that aligning the annotation protocol with the spatial structure of artifacts reduces annotator variance relative to global scalar ratings. 
The resulting masks also provide localized diagnostic feedback that may be useful for downstream model improvement~\cite{liu2025flow,fan2023reinforcement,wallace2024diffusion}.

\input{fig/visual_artifacts_figure}


\myparagraph{Convergence Analysis}
We further analyze how quickly each protocol stabilizes as annotators are added. 
For each annotator count $k$, we estimate reliability by repeatedly sampling random subsets of $k$ annotators and computing Krippendorff's $\alpha$. 
\new{We report the mean agreement and the $95\%$ range across subsets, shown as the shaded band, to quantify sensitivity to annotator choice.}
As shown in \Cref{fig:visual_artifacts_b}, Likert scoring remains unstable even with six annotators and exhibits a wide range across subsets. 
In contrast, brush-based annotation stabilizes with four annotators and maintains a consistently narrower band across $k$, indicating greater robustness to different subsets annotators.


\subsection{Text Rendering Accuracy}
\label{sec:text_rendering}
Text rendering is increasingly important for modern T2I systems. 
Earlier evaluations often used BQA, asking whether the rendered text exactly matches the prompt~\cite{chao2024dsg}. 
This binary protocol becomes overly coarse when models produce mostly legible text with minor spelling, or formatting errors. 

To analyze this effect, we sample 30 prompts tagged with \texttt{text\_rendering} and generate images using the same three models. 
Under BQA, annotators decide whether the rendered text is correct. 
As shown in \Cref{fig:text_rendering_b}, BQA yields moderate agreement, with $\alpha = 0.63\ [0.47,\,0.75]$, and an EDR of $28.75\%$. 
Likert scoring improves agreement to $\alpha = 0.81\ [0.71,\,0.86]$ and reduces EDR to $18.75\%$. 
However, both BQA and Likert scoring compress text rendering into a single global judgment, discarding information about which words are correct, missing, or incorrectly inserted.

\myparagraph{Word-Level Text Annotation}
To align annotation with the discrete structure of text, we use word-level annotation, as shown in \Cref{fig:text_rendering_a}. 
Annotators judge each expected word individually as correct or incorrect. 
Missing words receive a negative judgment on the corresponding word tile. 
To capture insertion errors, we add virtual \emph{gap tokens} between words and at sentence boundaries, which annotators mark when extra or repeated text appears. 
We compute the final accuracy as $\max(0, N_{\text{correct}} - N_{\text{incorrect\_gaps}}) / N_{\text{total\_words}}.$
\new{This formulation penalizes incorrect words, missing words, and unintended insertions while preserving partial correctness.}

\new{With word-level annotation, agreement increases to $\alpha = 0.89\ [0.79,\,0.94]$, and EDR drops to $3.9\%$.}
\new{When agreement is computed directly at the word and gap-token level, treating each tile as a binary item, agreement further increases to $\alpha = 0.96\ [0.95,\,0.97]$.}
These results show that decomposing text rendering into atomic units substantially reduces annotator variance compared to global judgments.


\myparagraph{Convergence Analysis}
We again measure agreement as a function of annotator count. 
For BQA and Likert protocols, agreement spans a wide band across different number of, reflecting instability in global judgments. 
In contrast, word-level annotation exhibit significantly tighter band across all $k$ indicating robustness to robust to the specific annotators sampled.

\input{fig/text_rendering_figure}

\subsection{Skills Requiring External References}
\label{sec:anchor_based}

Some skills require external or domain-specific knowledge, such as recognizing landmarks, artistic styles, or named entities. 
If annotators are unfamiliar with the reference, their judgments may become unreliable or default to abstention. 
To investigate this, we sample 30 prompts from \texttt{environment:landmark}, \texttt{style:artistic\_style}, and \texttt{named\_entities}, and generate images using the same three models. 
As a baseline, annotators answer BQA questions with an explicit \emph{Unsure} option. 
As shown in \Cref{fig:anchor_based_b}, this baseline yields an \emph{Unsure} rate of $63\%$, indicating that missing background knowledge is a major bottleneck.


\myparagraph{Anchor-Based Annotation}
To reduce dependence on annotator prior knowledge, we introduce \emph{Anchor} annotation. 
Given a validation question, an automated retrieval step presents the top-$k$ representative reference images for the queried entity or style alongside the generated image. 
Annotators then consult these visual anchors before responding, as illustrated in \Cref{fig:anchor_based_a}. 
We use $k=3$ to provide multiple references while keeping the interface compact.

Providing anchors reduces the \emph{Unsure} rate to $2.5\%$, suggesting that most abstentions arise from missing reference knowledge rather than inherent ambiguity in the generated images. 
\new{Agreement also increases from $\alpha = 0.47\ [0.25,\,0.68]$ to $\alpha = 0.57\ [0.47,\,0.66]$.}
However, annotators reported that strict binary decisions can be too rigid for stylistic or referential similarity judgments. 
We therefore also evaluate \emph{Anchor Likert}, which retains visual references but replaces binary responses with graded similarity ratings. 
\new{Anchor Likert further improves agreement to $\alpha = 0.64\ [0.54,\,0.72]$ while maintaining a low \emph{Unsure} rate.}


\myparagraph{Convergence Analysis}
As shown in \Cref{fig:anchor_based_b}, standard BQA does not stabilize with six annotators, indicating persistent sensitivity to annotator choice. 
In contrast, Anchor BQA and Anchor Likert reach stable agreement with approximately four annotators. 
\new{The wide shaded band for No-Anchor BQA at low $k$ shows high sensitivity to which annotators are sampled, whereas the narrower bands for anchor-conditioned variants indicate greater robustness.}

\new{
\subsection{Discussion}

\myparagraph{Number of Samples}
We initially sampled 20 prompts per skill category, but reliability statistics had not stabilized at that size. 
Increasing to 30 stabilized the estimates and clearly separated the confidence intervals between competing annotation mechanisms, indicating that 30 prompts suffice for comparing protocol reliability.

\myparagraph{Number of Annotators}
We used the same six professional annotators throughout to control for annotator effects. Convergence analyses show that reliability stabilizes at four to five annotators, suggesting that skill-aligned protocols can reduce annotation cost without sacrificing stability.}

\input{fig/anchor_based_figure}

%% file: fig/visual_artifacts_figure.tex
\begin{figure}[!tb]
    \centering

    \begin{subfigure}[c]{0.52\linewidth}
        \centering
        \includegraphics[
            trim={0 40 850 10},
            clip,
            width=\linewidth
        ]{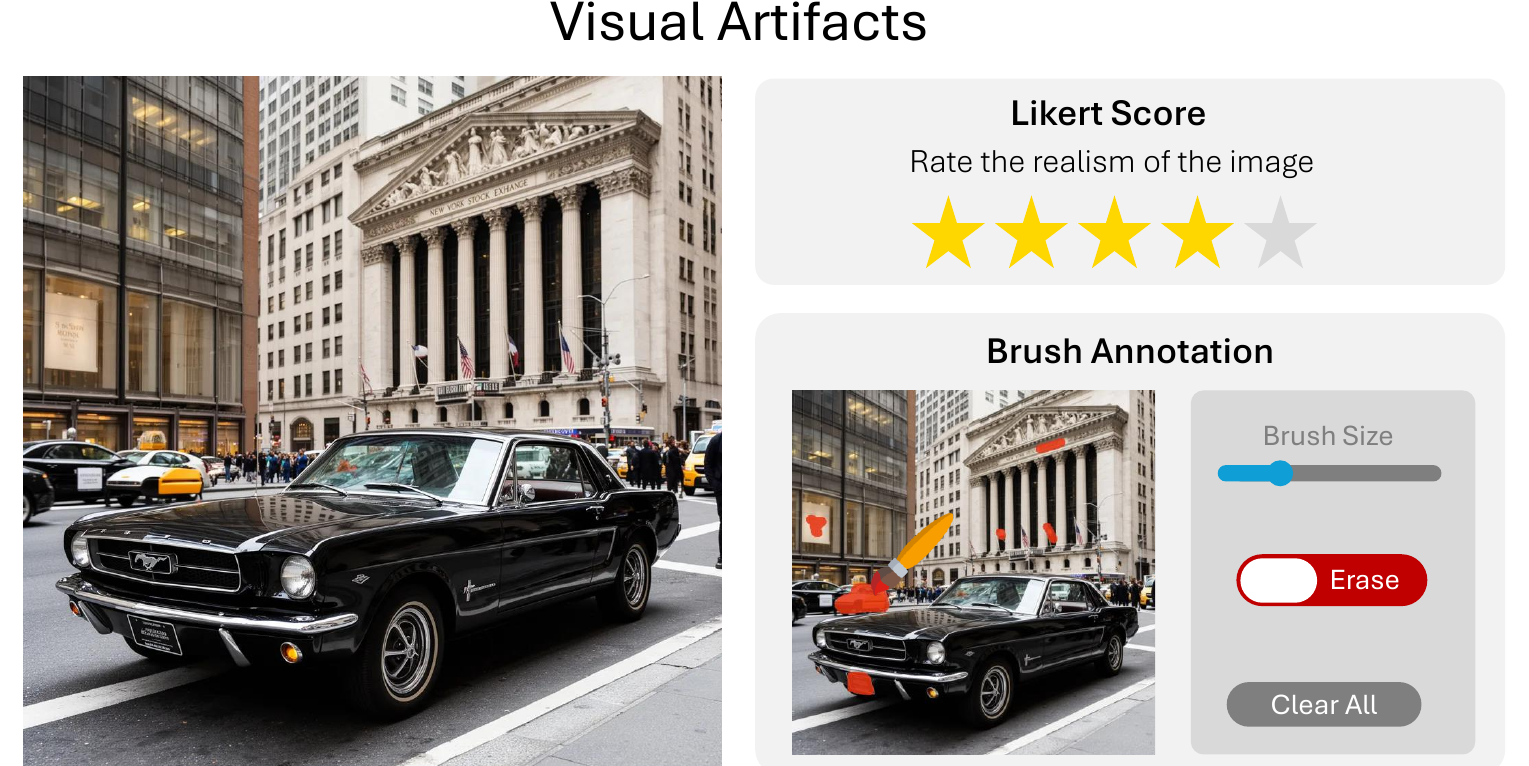}
        \caption{}
        \label{fig:visual_artifacts_a}
    \end{subfigure}
    \hfill
    \begin{subfigure}[c]{0.46\linewidth}
        \centering

        \scriptsize
        \setlength{\tabcolsep}{4pt}
        \renewcommand{\arraystretch}{1.2}

        \begin{tabular}{lcc}
            \toprule
             & \shortstack{\textbf{Inter-Annotator Agreement}\\[-1pt]
                           (Krippendorff's $\alpha$) $\uparrow$}
             & \shortstack{\textbf{EDR}\\[-1pt]
                           ($\geq 40\%$) $\downarrow$} \\
            \midrule
            Likert Score & $0.39\,[0.27,\,0.50]$ & $62.2\%$ \\
            Brush        & $\mathbf{0.82}\,[0.56,\,0.89]$ & $\mathbf{0.0\%}$ \\
            \bottomrule
        \end{tabular}

        \vspace{0.7em}

        \includegraphics[width=\linewidth]{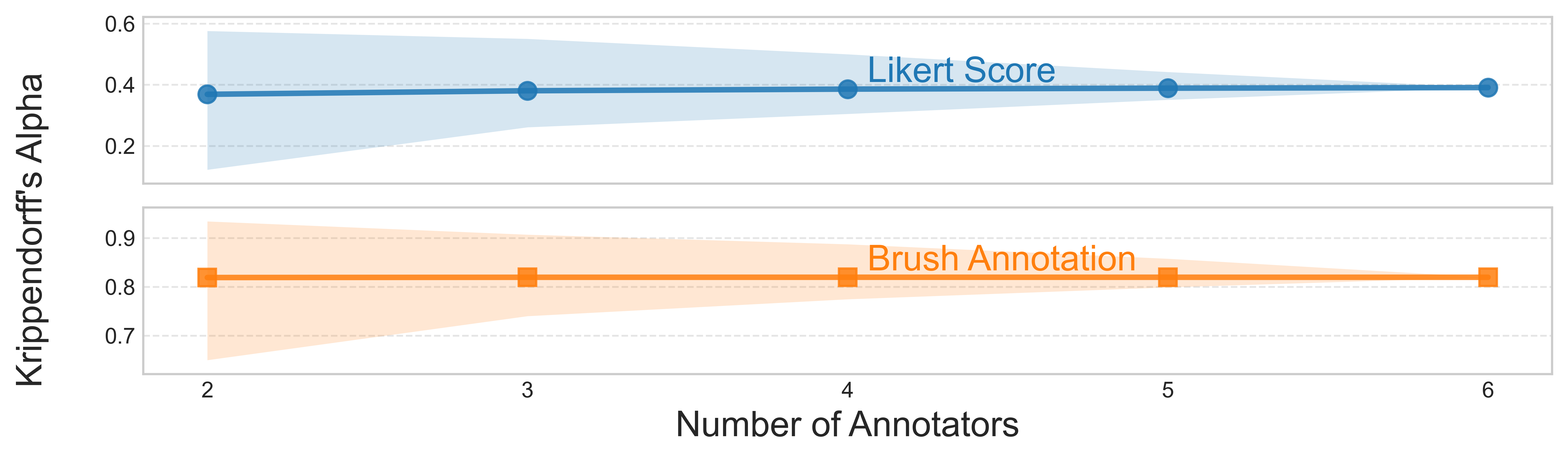}

        \caption{}
        \label{fig:visual_artifacts_b}
    \end{subfigure}

    \caption{Comparison between Likert scoring and brush-based annotation for
    visual artifacts in terms of inter-annotator agreement (Krippendorff's
    $\alpha$ with $95\%$ bootstrap CI), extreme disagreement rate, and
    convergence behavior.}
    \label{fig:visual_artifacts}
\end{figure}

%% file: fig/text_rendering_figure.tex
\begin{figure}[!t]
    \centering

    \begin{subfigure}[c]{0.52\linewidth}
        \centering
        \includegraphics[
            trim={0 40 750 10},
            clip,
            width=\linewidth
        ]{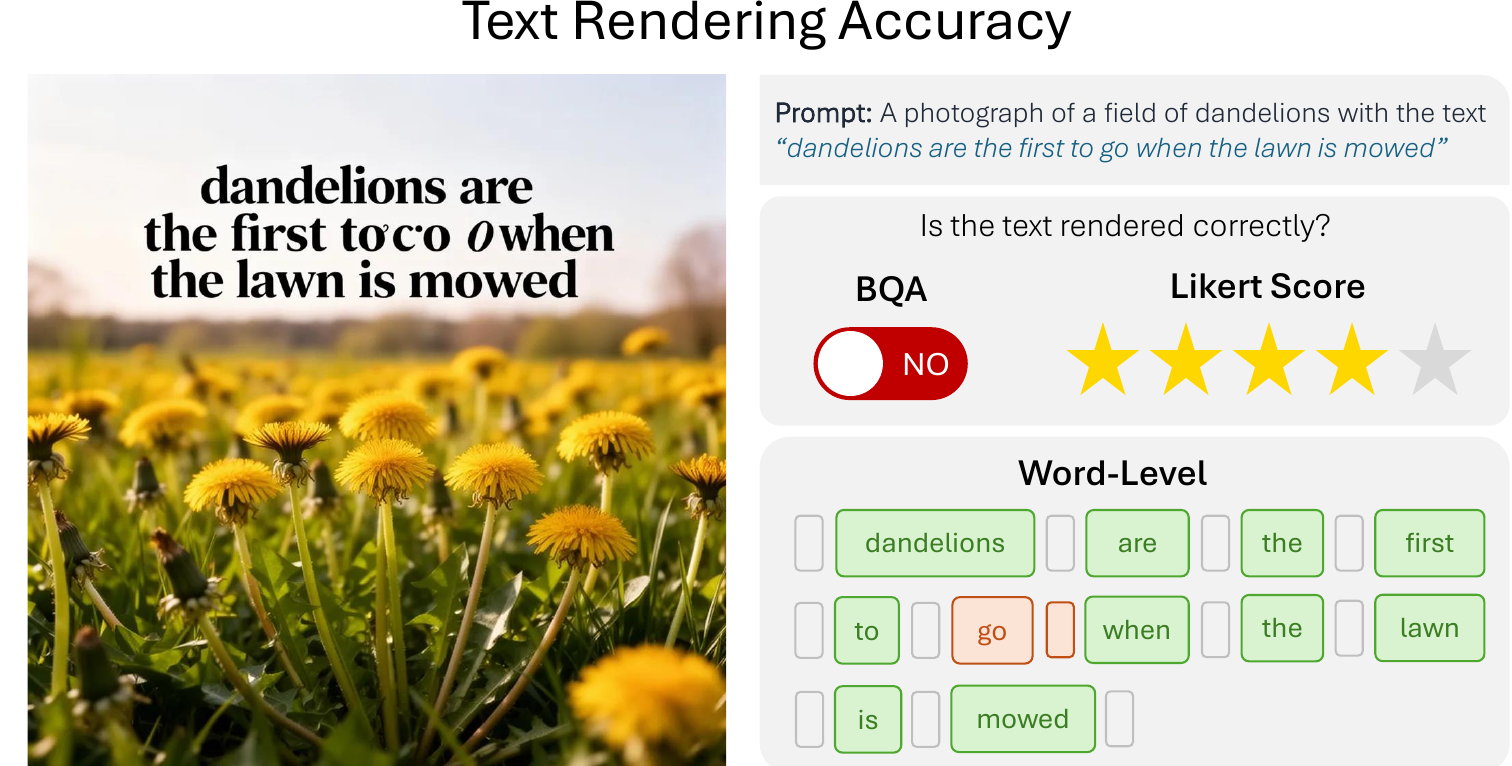}
        \caption{}
        \label{fig:text_rendering_a}
    \end{subfigure}
    \hfill
    \begin{subfigure}[c]{0.46\linewidth}
        \centering
        \tiny
        \setlength{\tabcolsep}{4pt}
        \renewcommand{\arraystretch}{1.2}

        \begin{tabular}{lcc}
            \toprule
             & \shortstack{\textbf{Inter-Annotator Agreement}\\[-1pt]
                           (Krippendorff's $\alpha$) $\uparrow$}
             & \shortstack{\textbf{EDR}\\[-1pt]
                           ($\geq 40\%$) $\downarrow$} \\
            \midrule
            BQA                  & $0.63 \ [0.47,\,0.75]$           & $28.75\%$        \\
            Likert Score         & $0.81 \ [0.71,\,0.86]$           & $18.75\%$        \\
            Word-Level (Whole)   & $0.89 \ [0.79,\,0.94]$           & $\mathbf{3.9\%}$ \\
            Word-Level (Per-Word)& $\mathbf{0.96} \ [0.95,\,0.97]$  & $4.5\%$          \\
            \bottomrule
        \end{tabular}

        \vspace{0.7em}

        \includegraphics[width=0.95\linewidth]{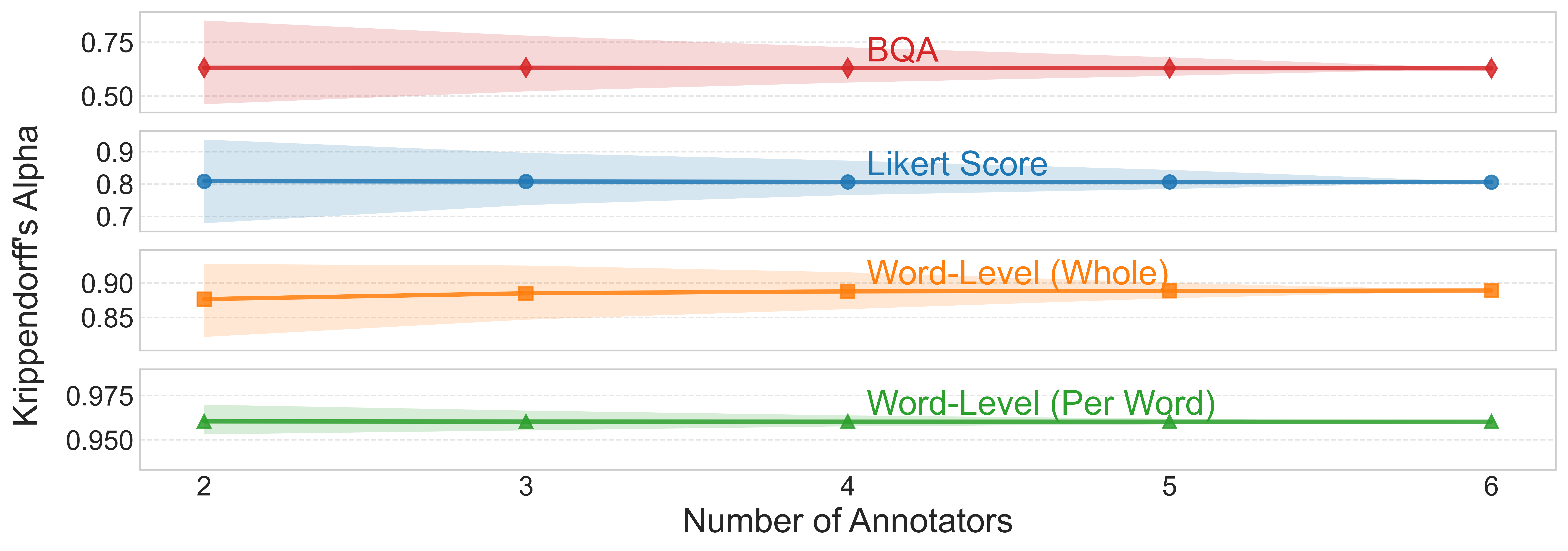}

        \caption{}
        \label{fig:text_rendering_b}
    \end{subfigure}

    \caption{Comparison between BQA, Likert, and word-level annotation for text
    rendering accuracy in terms of inter-annotator agreement (Krippendorff's
    $\alpha$ with $95\%$ bootstrap CI), extreme disagreement rate, and
    convergence behavior.}
    \label{fig:text_rendering}
\end{figure}

%% file: fig/anchor_based_figure.tex
\begin{figure}[!tb]
    \centering

    \begin{subfigure}[c]{0.48\linewidth}
        \centering
        \includegraphics[
            trim={0 30 930 10},
            clip,
            width=\linewidth
        ]{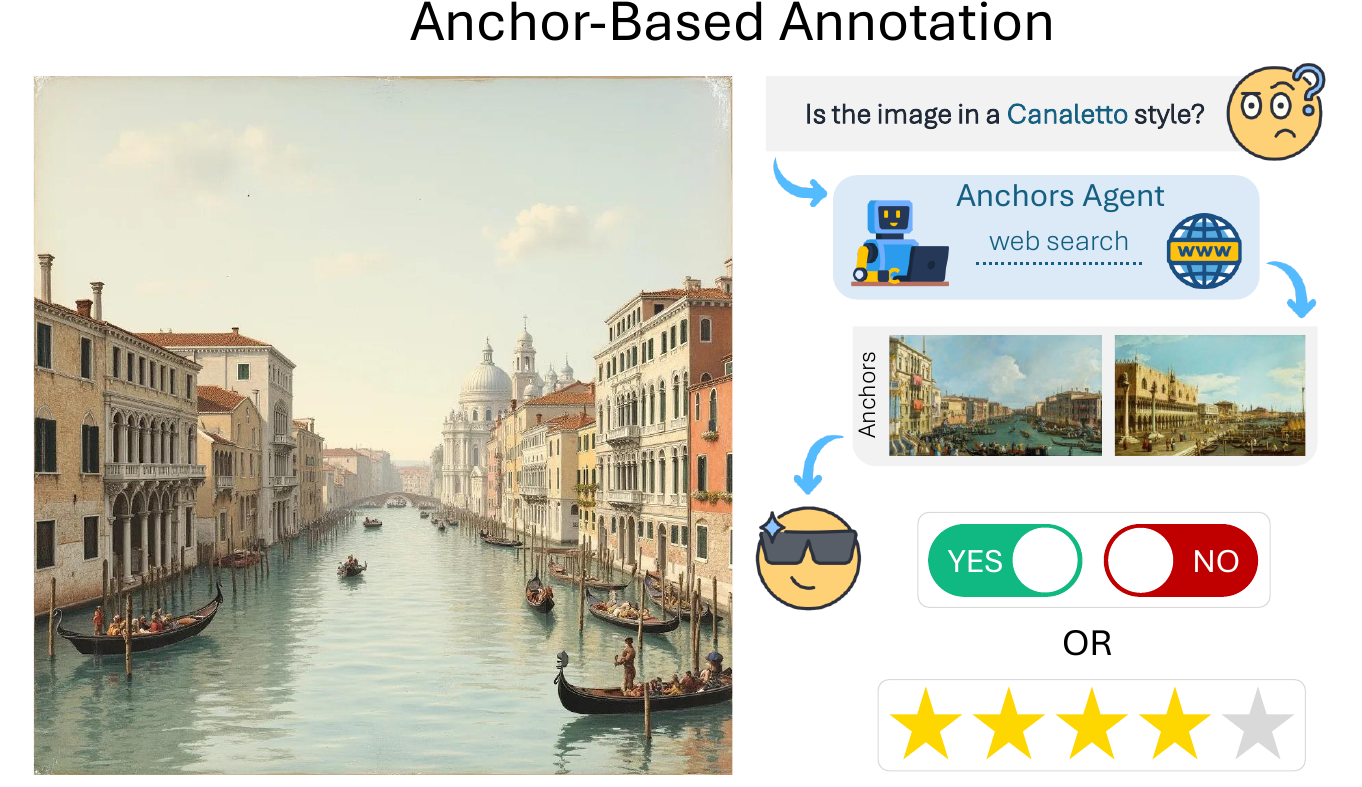}
        \caption{}
        \label{fig:anchor_based_a}
    \end{subfigure}
    \hfill
    \begin{subfigure}[c]{0.50\linewidth}
        \centering
        \scriptsize
        \setlength{\tabcolsep}{4pt}
        \renewcommand{\arraystretch}{1.2}

        \begin{tabular}{lcc}
            \toprule
             & \shortstack{\textbf{Inter-Annotator Agreement}\\[-1pt]
                           (Krippendorff's $\alpha$) $\uparrow$}
             & \shortstack{\textbf{Unsure Rate}\\[-1pt]
                           (\%) $\downarrow$} \\
            \midrule
            No Anchor BQA  & $0.47 \ [0.25,\,0.68]$           & $63.0\%$         \\
            Anchor BQA     & $0.57 \ [0.47,\,0.66]$           & $\mathbf{2.5\%}$ \\
            Anchor Likert  & $\mathbf{0.64} \ [0.54,\,0.72]$  & $\mathbf{2.5\%}$ \\
            \bottomrule
        \end{tabular}

        \vspace{0.7em}

        \includegraphics[width=\linewidth]{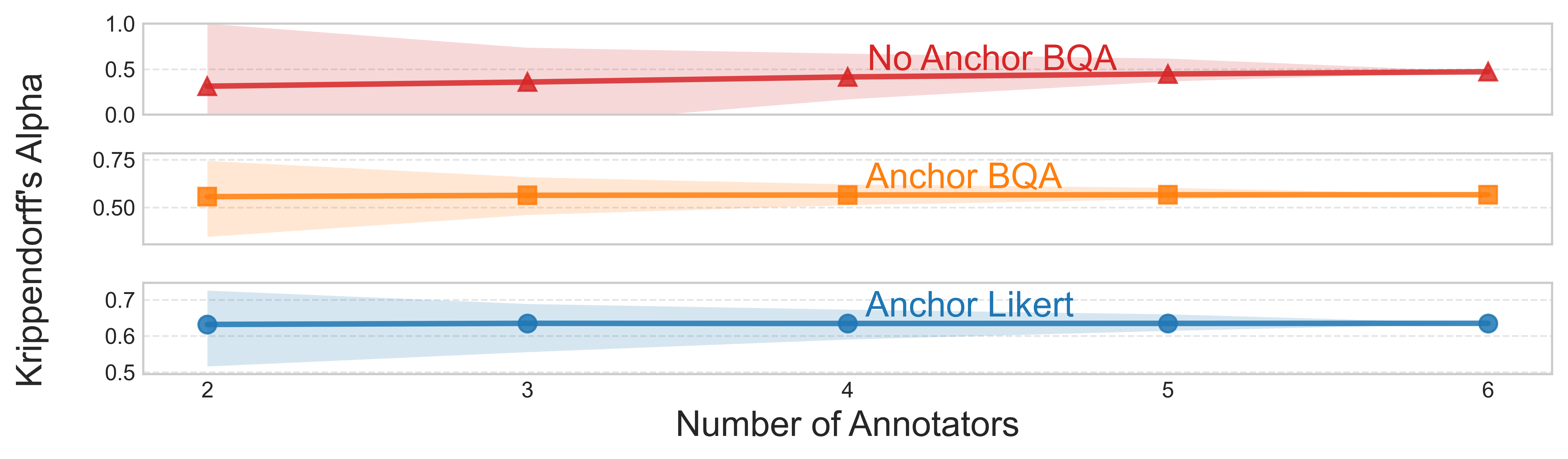}

        \caption{}
        \label{fig:anchor_based_b}
    \end{subfigure}

    \caption{Comparison between standard BQA and Anchor-based protocols in terms of
    inter-annotator agreement (Krippendorff's $\alpha$ with $95\%$ bootstrap CI),
    unsure rate, and convergence behavior.}
    \label{fig:anchor_based}
\end{figure}

%% file: sec/5_full_eval.tex
\section{Full Evaluation Suite}

Based on the controlled comparisons in the previous section, we construct a unified evaluation protocol that assigns each skill the annotation strategy with the strongest reliability profile. 
The final suite uses brush-based annotation for visual artifacts, word-level annotation with gap tokens for text rendering, Anchor Likert for reference-dependent skills, BQA for strictly factual skills, and Likert scoring for overall aesthetic quality.

For full evaluation, we sample 43 prompts from the taxonomy in \Cref{sec:taxonomy}, ensuring coverage across all skills and sub-skills. 
We evaluate two proprietary models, Nano-Banana (Gemini-Flash-2.5)~\cite{nanobanana2025} and Wan-2.5-preview~\cite{wan25}, and two open-source models, FLUX.2-dev~\cite{flux-2-2025} and Z-Image-Turbo~\cite{cai2025z}. 
To compare heterogeneous outputs, all non-binary scores, including Likert ratings, brush-based artifact ratios, and word-level accuracy, are normalized to $[0,1]$.

\begin{table*}[t]
\centering
\scriptsize
\setlength{\tabcolsep}{4pt}
\renewcommand{\arraystretch}{1.2}

\begin{tabularx}{\textwidth}{
l
*{9}{>{\centering\arraybackslash}X}
}
\toprule
\textbf{Model}
& \faCube 
& \faSlidersH 
& \faRunning 
& \faProjectDiagram 
& \faBalanceScale 
& \faLightbulb 
& \faCloudSun 
& \faEye 
& \faCamera \\
\midrule

Nano-Banana~\cite{nanobanana2025} 
& 0.979 & 0.900 & 0.773 & \textbf{0.978} & 0.950 
& \textbf{0.955} & \textbf{1.000} & 0.938 & 0.688 \\

Wan-2.5-preview~\cite{wan25} 
& \textbf{0.992} & 0.920 & 0.793 & 0.960 & \textbf{0.976} 
& 0.875 & \textbf{1.000} & \textbf{1.000} & 0.778 \\

FLUX.2-dev~\cite{flux-2-2025} 
& 0.934 & \textbf{0.936} & \textbf{0.831} & 0.880 & 0.743 
& 0.907 & 0.933 & 0.688 & 0.882 \\

Z-Image-Turbo~\cite{cai2025z} 
& 0.913 & 0.827 & 0.730 & 0.705 & 0.561 
& 0.625 & 0.935 & 0.875 & \textbf{0.905} \\

\bottomrule
\end{tabularx}

\begin{tabularx}{\textwidth}{
l
*{9}{>{\centering\arraybackslash}X}
>{\centering\arraybackslash}X
}
\textbf{Model}
& \faSmile 
& \faLanguage 
& \faClock 
& \faMountain 
& \faPaintBrush 
& \faIdBadge 
& \faFont 
& \faExclamationTriangle 
& \faStar
& \textbf{Avg} \\
\midrule

Nano-Banana~\cite{nanobanana2025} 
& \textbf{1.000} & \textbf{1.000} & 0.967 & 0.901 
& \textbf{0.875} & 0.639 & 0.930 & \textbf{0.998} & \textbf{0.879}
& \textbf{0.908} \\

Wan-2.5-preview~\cite{wan25} 
& \textbf{1.000} & 0.743 & \textbf{0.969} & 0.880 
& 0.783 & 0.615 & \textbf{0.980} & 0.997 & 0.850
& 0.895 \\

FLUX.2-dev~\cite{flux-2-2025} 
& 0.867 & 0.848 & 0.933 & \textbf{0.908} 
& 0.817 & \textbf{0.659} & 0.941 & 0.996 & 0.824
& 0.863 \\

Z-Image-Turbo~\cite{cai2025z} 
& \textbf{1.000} & 0.176 & 0.767 & 0.835 
& 0.691 & 0.485 & 0.877 & {0.995} & 0.796
& 0.761 \\

\bottomrule
\end{tabularx}

\vspace{6pt}
\caption{
Per-skill comparison of diffusion models under the full skill-adaptive evaluation protocol. 
Bold indicates best performance per skill. 
Icons correspond to: 
\faCube\ Entities, 
\faSlidersH\ Attributes, 
\faRunning\ Action, 
\faProjectDiagram\ Spatial Arrangement, 
\faBalanceScale\ Relative Comparison, 
\faLightbulb\ Lighting, 
\faCloudSun\ Weather, 
\faEye\ Viewpoint, 
\faCamera\ Camera Control, 
\faSmile\ Mood/Feeling, 
\faLanguage\ Language Complexity, 
\faClock\ Time, 
\faMountain\ Environment/Landmark, 
\faPaintBrush\ Style, 
\faIdBadge\ Named Entities, 
\faFont\ Text Rendering, 
\faExclamationTriangle\ Visual Artifacts, 
and \faStar\ Aesthetic Quality.
}
\label{tab:full_evaluation}
\end{table*}


\Cref{tab:full_evaluation} summarizes per-skill performance across models. 
Nano-Banana obtains the highest average score, followed by Wan-2.5-preview, but aggregate rankings hide important skill-level differences.

\myparagraph{Visual Artifacts}
Brush-based artifact scores are nearly saturated across models, suggesting that recent T2I systems have largely closed the gap in low-level artifact suppression. 
The small margins highlight the value of spatial annotation, since global ratings could obscure localized differences.

\myparagraph{Named Entities}
All models perform poorly on named entities, making this one of the most challenging evaluated skills. 
This likely reflects the difficulty of grounding generation in a large and open-ended reference space, motivating stronger retrieval or grounding mechanisms.

\myparagraph{Language Complexity and Relational Reasoning}
Language complexity shows the largest variance across models. 
Z-Image-Turbo scores $0.176$, substantially below the other systems, and also performs weakly on spatial arrangement and relative comparison. 
This suggests that the model trades advanced prompt fidelity for efficiency. 
In contrast, Nano-Banana achieves perfect performance on language complexity, consistent with stronger instruction following from its backbone LLM.


\subsection{Score Convergence Across Skills}

\begin{figure}[!t]
    \centering
    \includegraphics[width=\linewidth]{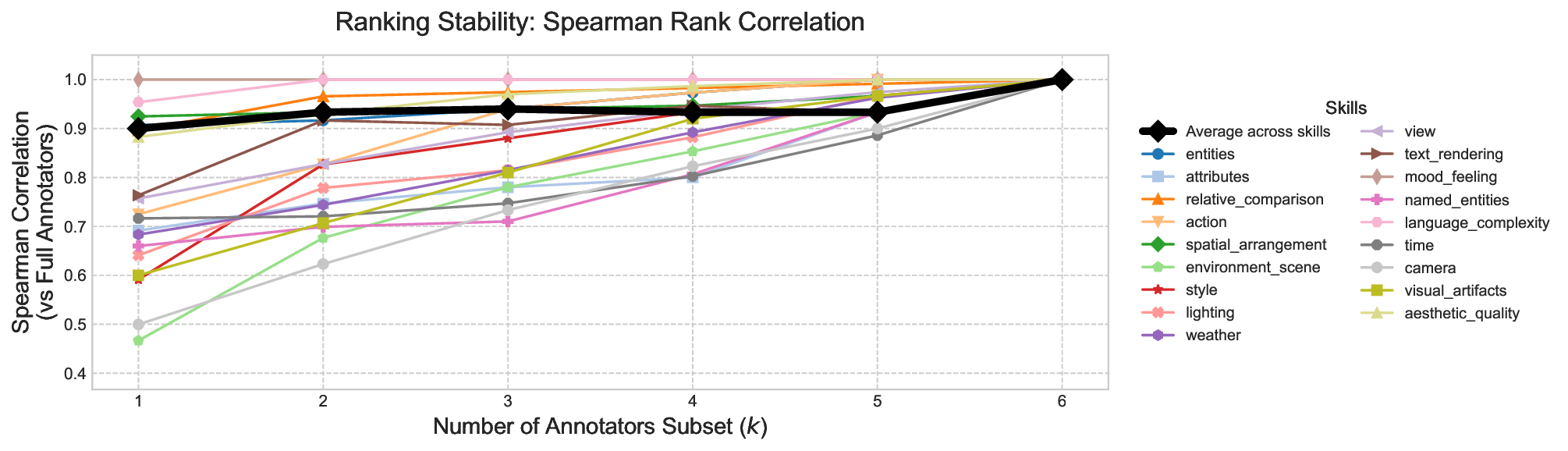}
    \caption{Score convergence under annotator subsampling. 
    Spearman correlation between subset-based model scores and full-annotator scores is shown as a function of annotator count. }
    \label{fig:full_eval_stability}
\end{figure}

We evaluate score stability by measuring how model rankings change under annotator subsampling. 
For each annotator count $k$, we sample 50 random subsets of $k$ annotators, recompute model scores, and measure Spearman rank correlation with the full-annotator scores.
\Cref{fig:full_eval_stability} shows that rankings remain stable across annotator counts. 
Aggregate correlations are consistently high, indicating that the overall model ordering is robust to annotator subsampling. 
At the skill level, convergence varies slightly: \texttt{relative\_comparison} and \texttt{language\_complexity} stabilize rapidly, whereas \texttt{environment\_scene} and \texttt{style} require more annotators due to greater subjectivity and reliance on reference anchors. 
Nevertheless, all skills exceed a Spearman correlation of $0.8$ by four annotators, showing that our evalution suite yields stable model rankings with a modest annotator pool.

%% file: sec/6_automated.tex
\section{Automatic Evaluation with LLM Agents}

To test fully automatic evaluation, we replicate our skill-adaptive protocol with a lightweight agentic framework in which separate agents execute each skill-specific procedure.
We evaluate two open-source VLMs, Molmo2-8B~\cite{clark2026molmo2} and Qwen-3.5-9B~\cite{qwen35}, and the proprietary ChatGPT-5.
Open-source models were deployed on a single A100 GPU (80GB VRAM).
For BQA, the agent answers validation questions with \textit{yes}/\textit{no}.
For reference-dependent skills, it receives a collage of the generated image and two labeled anchors and assigns a graded similarity score using the human Anchor Likert rubric.
For word-level text rendering, it labels each expected word and inter-word gap individually, and per-image accuracy is computed as in the human protocol.
Since LLMs cannot brush spatial artifacts, we use a pretrained artifact detector~\cite{zhang2023perceptual} for artifact masks.

\begin{figure}[!t]
    \centering
    \includegraphics[width=\linewidth]{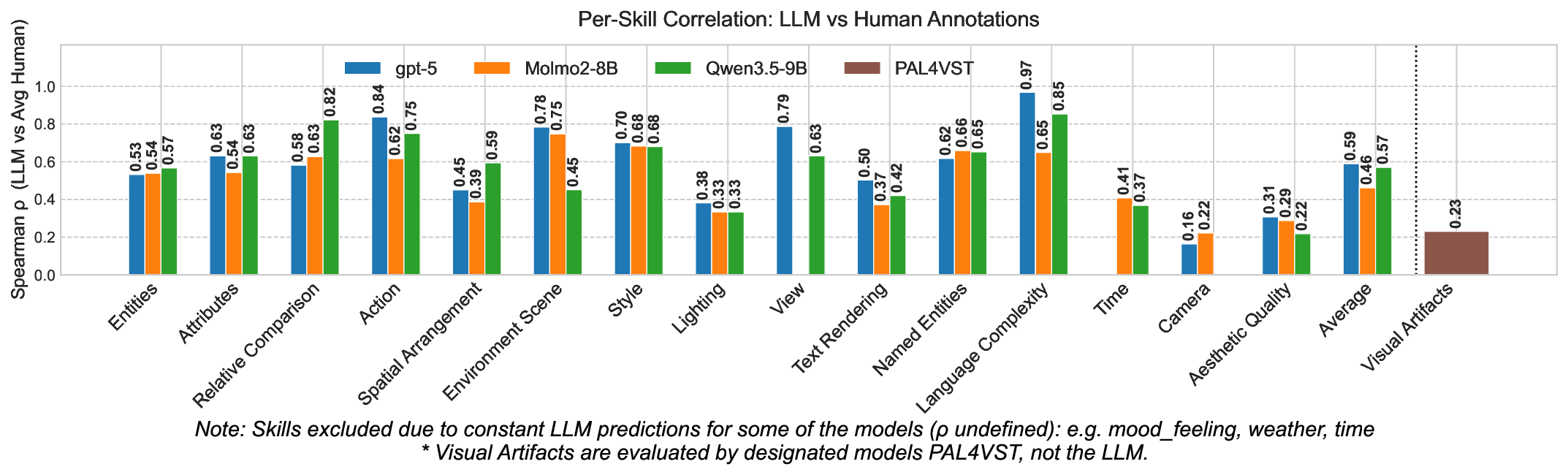}
    \caption{Spearman correlation between LLM-based automatic evaluation and human evaluation across skills. Structured semantic skills show higher alignment, while camera-related attributes and aesthetic quality exhibit lower correlation.}
    \label{fig:llm_alignment}
\end{figure}

\Cref{fig:llm_alignment} shows that most structured semantic skills achieve moderate to strong correlation with humans, indicating the protocol transfers reasonably well to automatic assessment. 
Notably, Qwen-3.5 performs comparably to ChatGPT-5 despite being orders of magnitude smaller.
Aesthetic quality shows substantially lower alignment, suggesting generic LLM judgment is insufficient and may benefit from specialized predictors (\eg ImageReward~\cite{xu2023imagereward}).
Camera-related attributes also correlate poorly, indicating that humans and VLMs interpret camera properties differently.
Artifact detection with PAL4VST shows relatively low agreement, likely because it was trained on earlier diffusion models and requires adaptation to newer ones.
Among structured skills, \texttt{entities} and \texttt{spatial\_arrangement} reveal limits in counting and multi-object reasoning, while word-level text rendering achieves only moderate alignment, indicating the need for specialized model or human inspection.
Overall, the protocol generalizes across model families, but certain perceptual and compositional skills remain intrinsically difficult to evaluate automatically.

%% file: sec/7_conclusion.tex
\section{Conclusion and Future Work}

We studied \emph{skill-aligned annotation} for text-to-image evaluation, where protocols are matched to the structure of each skill.
In controlled comparisons, skill-aligned protocols produced more reliable and stable signals than uniform mechanisms even with a modest annotator pool, showing that careful evaluation design can improve reliability without scaling annotation.
We also instantiated the protocol in an automated pipeline that aligns strongly with human judgments on several skills, indicating its potential to partially automate large-scale T2I evaluation while preserving fine-grained feedback.

\myparagraph{Future Work}
Our study targets skills where uniform annotation is likely unreliable: artistic style, named entities, landmarks, text rendering, and visual artifacts.
However, the convergence analysis in \Cref{fig:full_eval_stability} and the human--LLM alignment in \Cref{fig:llm_alignment} indicate that weather, camera, and mood/feeling also converge slowly across annotators and warrant skill-aligned protocols.
Annotation mechanisms should also be refined for failures not fully captured by the current design, such as missing objects or object parts in artifact annotation.
More broadly, the same methodology can extend to axes such as safety, cultural alignment, and diversity, where reliable protocols remain underexplored.
We hope this work encourages treating annotation design as a central component of evaluation methodology, so benchmarks measure intended capabilities rather than artifacts of the annotation process.

%% file: sec/appendix.tex


\section{Details on Skills Taxonomy}
\label{sec:skill_details}
We provide additional examples illustrating the proposed skill–subskill taxonomy in \Cref{tab:skill_examples}. The taxonomy was iteratively refined until it consistently captured the diverse skills in all textual prompts present in the Gecko benchmark \cite{wiles2025revisiting}.
While this version covers the skills observed in Gecko, the framework is designed to remain extensible. 
If new skills emerge, the taxonomy can be readily updated by modifying the tagging prompt to incorporate the additional skill categories.

\small
\setlength{\tabcolsep}{4pt}
\renewcommand{\arraystretch}{1.3}

\begin{longtable}{
>{\raggedright\arraybackslash}m{1.8cm}
>{\raggedright\arraybackslash}m{1.8cm}
>{\raggedright\arraybackslash}p{4.7cm}
>{\raggedright\arraybackslash}p{4.7cm}
}

\caption{Example prompts and validation questions for each evaluation skill and sub-skill. Questions follow the annotation strategies.}
\label{tab:skill_examples} \\

\toprule
\textbf{Skill} & \textbf{Sub-skill} & \textbf{Example Prompt} & \textbf{Validation Question} \\
\midrule
\endfirsthead

\toprule
\textbf{Skill} & \textbf{Sub-skill} & \textbf{Example Prompt} & \textbf{Validation Question} \\
\midrule
\endhead

\bottomrule
\endfoot

\faCube\ Entities & Singular & A photo of a \underline{cat}  & Is there exactly one cat in the image? \\

 & Count & \underline{Three apples} on a white plate & Are there three apples? \\

 & Uncountable & A glass filled with \underline{milk} & Does the image contain milk in the glass? \\

\midrule

\faSlidersH\ Attributes & Color & A \underline{blue} bicycle & Is the bicycle blue? \\

 & Texture & A \underline{rough} stone sculpture & Does the sculpture have a rough texture? \\

 & Material & A \underline{wooden} chair in a living room & Is the chair made of wood? \\

 & Shape & A perfectly \underline{round} mirror hanging on a wall & Is the mirror round? \\

 & Scale & A \underline{tiny} elephant playing & Is the elephant tiny?\\

\midrule

\faRunning\ Action & Standard & A dog \underline{running} across a field & Is the dog running? \\

 & Unusual & A dog \underline{riding a skateboard} & Is the dog riding a skateboard? \\

 & Pose & A gymnast performing a \underline{handstand} & Is the gymnast in a handstand pose? \\

\midrule

\faProjectDiagram\ Spatial Arrangement & -- & A cup placed \underline{on top of} a book & Is the cup on top of the book? \\

\midrule

\faBalanceScale\ Comparison & Scale & A \underline{large} elephant next to a \underline{small} dog & Is the elephant larger than the dog? \\

 & Tone & A \underline{darker} cloud beside a \underline{lighter} cloud & Is the left cloud darker than the right cloud? \\

 & Distance & A tree \underline{closer to the camera} than a house & Is the tree closer to the camera than the house? \\

 & Count & \underline{More birds than airplanes} in the sky & Are there more birds than airplanes? \\

\midrule

\faLightbulb\ Lighting & -- & A portrait illuminated by \underline{soft studio lighting} & Is the scene softly lit? \\

\faCloudSun\ Weather & -- & A \underline{snowy} street during a snowstorm & Is it snowing in the scene? \\

\midrule

\faEye\ View & -- & A \underline{top-down view} of a dining table with food & Is the scene viewed from above? \\

\midrule

\faCamera\ Camera & -- & A \underline{macro close-up} shot of a butterfly & Is this a close-up (macro) shot? \\

\midrule

\faSmile\ Mood / Feeling & -- & A \underline{gloomy} abandoned house under dark clouds & Does the image convey a gloomy mood? \\

\midrule

\faLanguage\ Language Complexity & Negation & A street with cars but \underline{no pedestrians} & Are there no pedestrians in the scene? \\

 & Color Stroop & The word \underline{“RED”} written in \underline{blue} ink & Is the word RED written in blue color? \\

\midrule

\faClock\ Time & Time of Day & A beach at \underline{sunset} with orange sky & Is the scene at sunset? \\

 & Season & Trees covered in \underline{autumn leaves} & Does the scene depict autumn? \\

 & Era & A \underline{medieval knight} beside a castle & Does the scene depict a medieval era? \\

\midrule

\faMountain\ Environment / Scene & General & A \underline{tropical beach} with palm trees & Does the image resemble a tropical beach scene? \\

 & Landmark & The \underline{Eiffel Tower} at night & Does the image depict the Eiffel Tower? \\

\midrule

\faPaintBrush\ Style & Artistic Style & A portrait in the style of \underline{Van Gogh} & How similar is the style to Van Gogh?  \\

 & Visual Medium & A landscape painting in on \underline{vase} & Is the painting on a vase? \\

\midrule

\faIdBadge\ Named Entities & Character & \underline{Spider-Man} swinging between buildings & Does the image depict Spider-Man? \\

 & Vehicle & A \underline{Tesla Cybertruck} parked on a street & Is the vehicle a Tesla Cybertruck? \\

 & Product & A \underline{Coca-Cola} can on a wooden table & Is the product Coca-Cola? \\

 & Artwork & The \underline{Mona Lisa} displayed in a museum & Does the image depict the Mona Lisa? \\

\midrule

\faFont\ Text Rendering & Accuracy & A sign that reads \underline{“OPEN 24 HOURS”} & Does the image have text “OPEN 24 HOURS”? \\

 & Style & The word "Hello" in \underline{cursive} style & Is the word "Hello" in cursive style?? \\

 & Numerical & A scoreboard showing \underline{3–1} & Does the scoreboard show numbers "3-1"? \\

 & Position & A label “EXIT” \underline{above a door} & Is the word "EXIT" above the door? \\

\end{longtable}
\normalsize

\section{Skill Tagging Agent}
\label{sec:tagging_app}

\begin{figure}[!t]
    \centering
    \includegraphics[width=\linewidth]{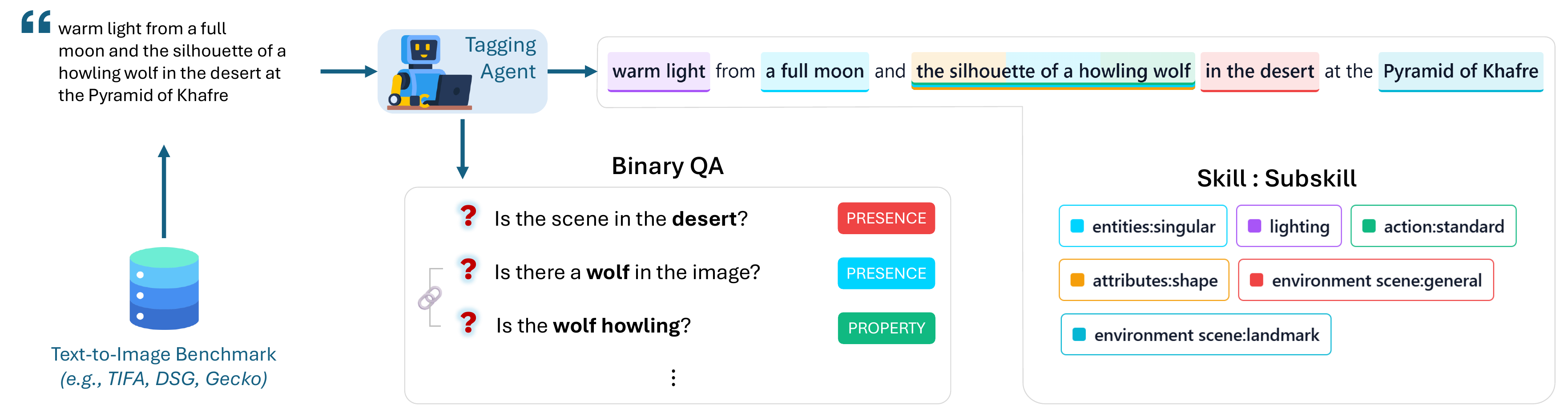}
    \caption{Our automated pipeline to tag Text-to-Image generation prompts with relevant evaluation skills and corresponding binary evaluation questions.}
    \label{fig:tagging}
\end{figure}

We design a single agent that simultaneously performs three tasks: 
(1) tagging different parts of the prompt with the relevant skills, 
(2) generating validation questions, and 
(3) identifying dependency relationships between questions.
While earlier approaches often relied on multiple agents or separate LLM calls for these steps, we find that recent LLMs can reliably perform them within a single prompt. 
In practice, generating the tags, questions, and dependencies jointly leads to more consistent and task-aligned outputs compared to performing each step independently.
An overview of the tagging pipeline is shown in \Cref{fig:tagging}.
The full prompt is provided in the source code.

\begin{figure}[!t]
    \centering
    \includegraphics[width=\linewidth]{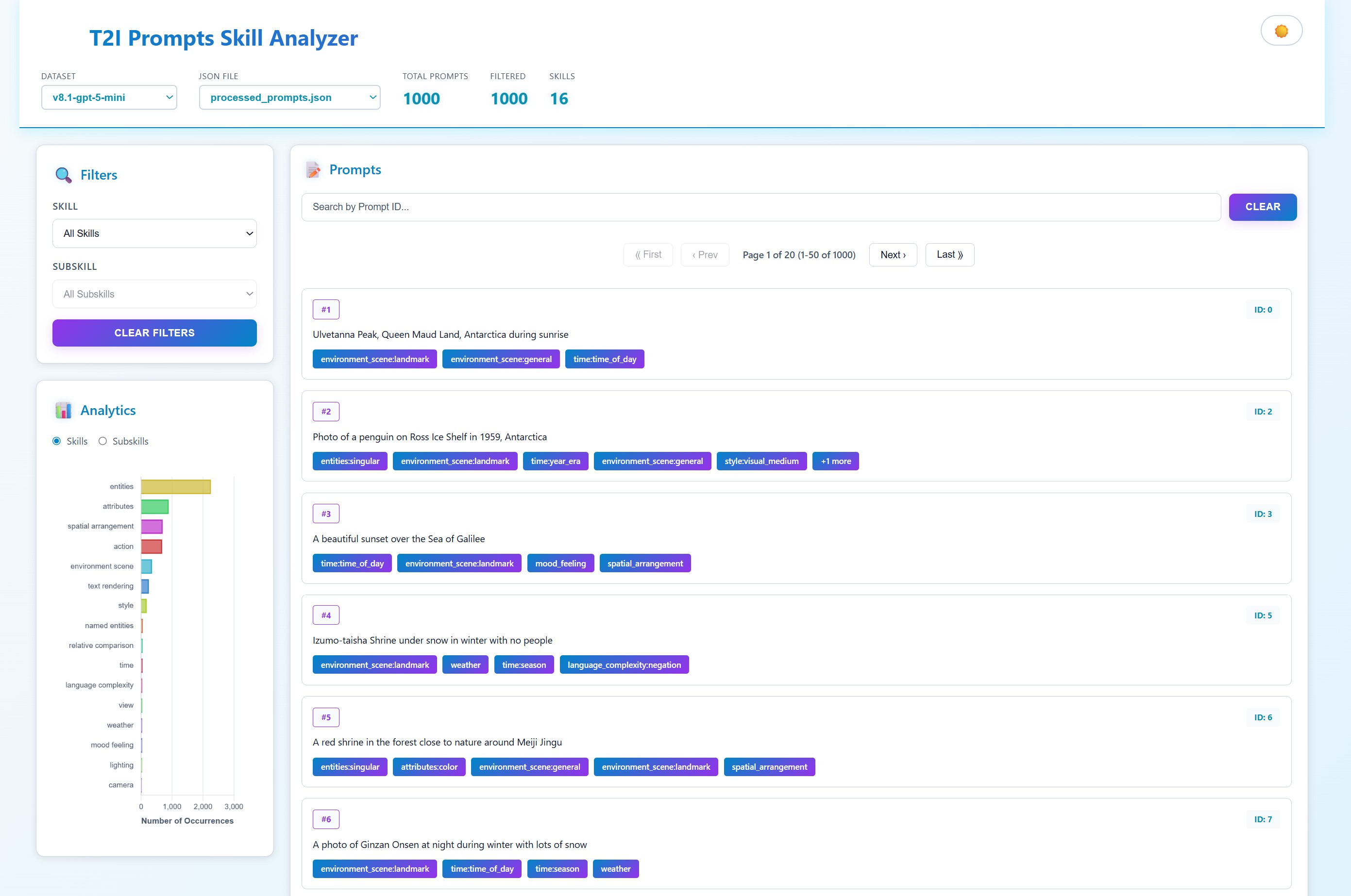}
    \caption{Overview of the prompt analysis interface. The application summarizes the distribution of skills and sub-skills across the prompts of a dataset and allows users to browse individual prompts.}
    \label{fig:prompt_app}
\end{figure}

\begin{figure}[!t]
    \centering
    \includegraphics[width=\linewidth]{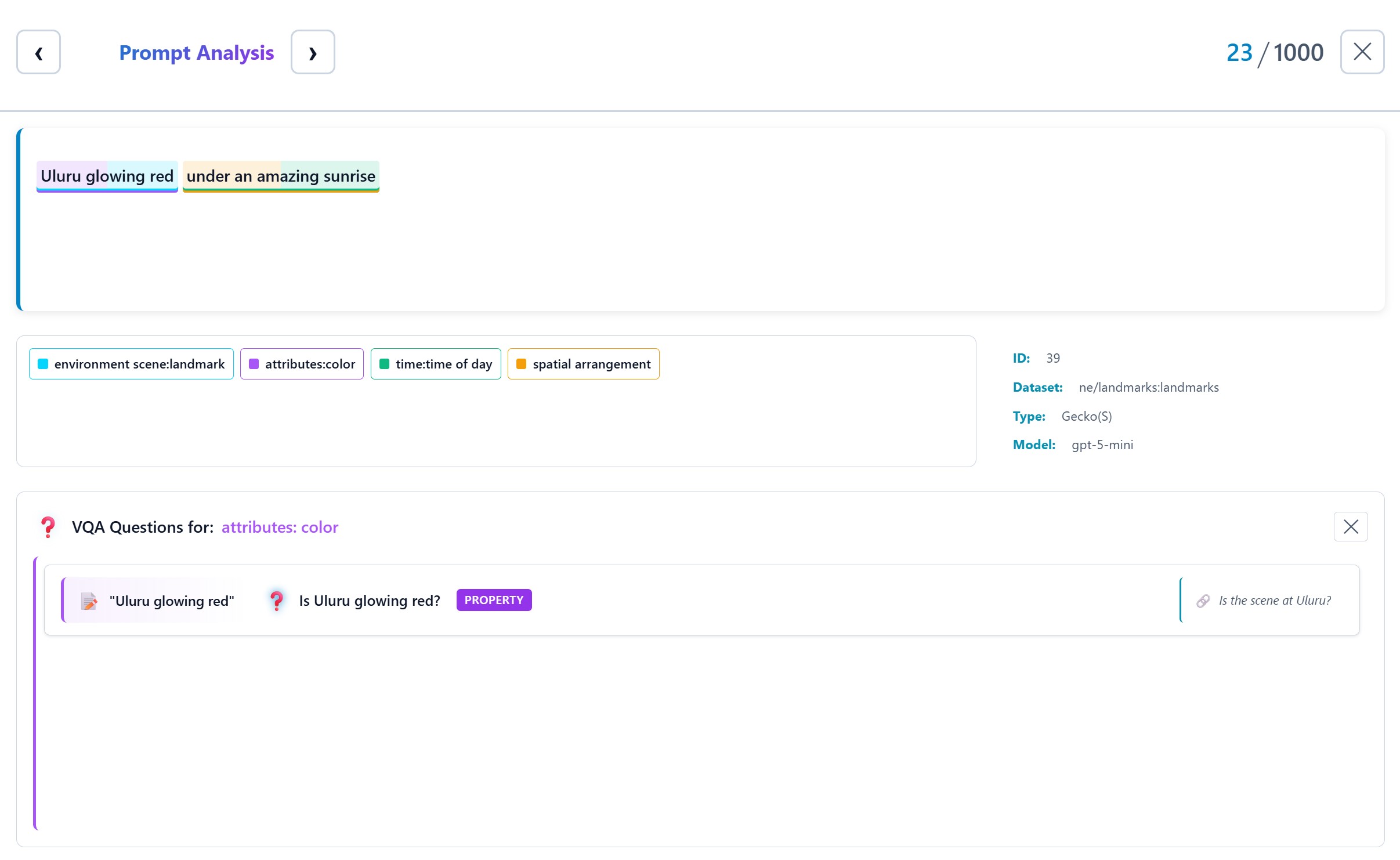}
    \caption{Detailed view for a selected prompt showing the automatically tagged skills, generated validation questions, and dependency relationships between questions.}
    \label{fig:prompt_app_details}
\end{figure}

\myparagraph{Prompt Analysis Application}
To facilitate inspection of prompts and their automatically tagged skills and validation questions, we developed a lightweight web application. 
An overview of the interface is shown in \Cref{fig:prompt_app}. The application summarizes the distribution of skills and sub-skills across all prompts in a dataset. 
When a specific prompt is selected, the interface displays detailed information including the tagged skills, the generated validation questions, and the dependency relationships between questions, as illustrated in \Cref{fig:prompt_app_details}.
A video demo of the application is provided in the the supplementary material.
The source code for the application and the tagged prompts will be made publicly available.

\section{Annotation Interface}

\begin{figure}[!t]
    \centering
    \includegraphics[width=\linewidth]{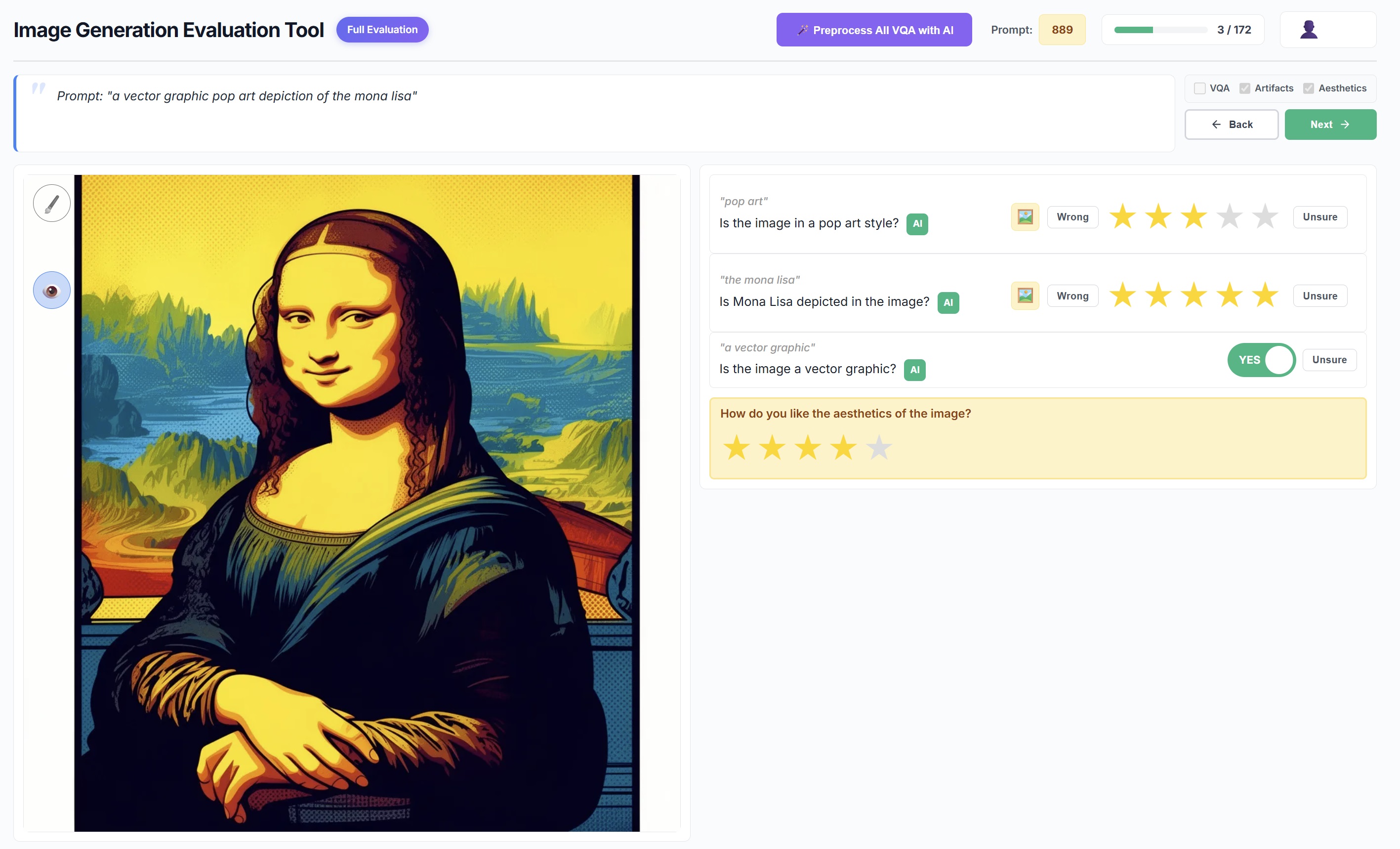}
    \caption{Web-based annotation interface used for skill-specific evaluation. Prompts, generated images, and corresponding annotation controls are displayed together according to the configured evaluation protocol.}
    \label{fig:eval_app}
\end{figure}

We developed a web-based annotation interface to streamline the evaluation workflow. 
A screenshot of the interface is shown in \Cref{fig:eval_app}. 
The application loads tagged prompts according to a predefined task configuration specifying the annotation strategy for different skills. 
An example of task configurations is shown below:
\begin{lstlisting}[style=jsonstyle]
  {
    "id": "text_per_word",
    "name": "Text Per Word",
    "enable_bqa_ai": false,
    "shuffle_images": true,
    "annotations": [
      "text_per_word"
    ],
    "dataset_version": "v8.1-gpt-5-mini",
    "prompts_file": "text_rendering_collection.json",
    "models": [
      "z-image",
      "flux2-dev",
      "flux1-dev"
    ]
  }
\end{lstlisting}

Generated images are then displayed together with their evaluation questions and corresponding annotation controls. 
For Anchor-based annotations, an image icon appears next to the question and reveals the anchor reference images when hovered over.
If a child question depends on a parent question, the child is disabled if the answer to the parent is negative.

To accelerate annotation, the interface optionally initializes answers using an LLM. These suggestions are explicitly labeled as \textit{AI} to ensure that annotators remain aware of their origin.
For artifact annotation, users can mark defective regions directly on the image. When selecting the artifact tool, the image becomes a drawing canvas that allows annotators to brush over areas containing visual artifacts, as illustrated in \Cref{fig:eval_app_artifacts}.
A video demo of the annotation app is provided in the supplementary material.
To make the evaluation of T2I generation models accessible, we will make the source code of the application publicly available.

\begin{figure}[!t]
    \centering
    \includegraphics[width=0.6\linewidth]{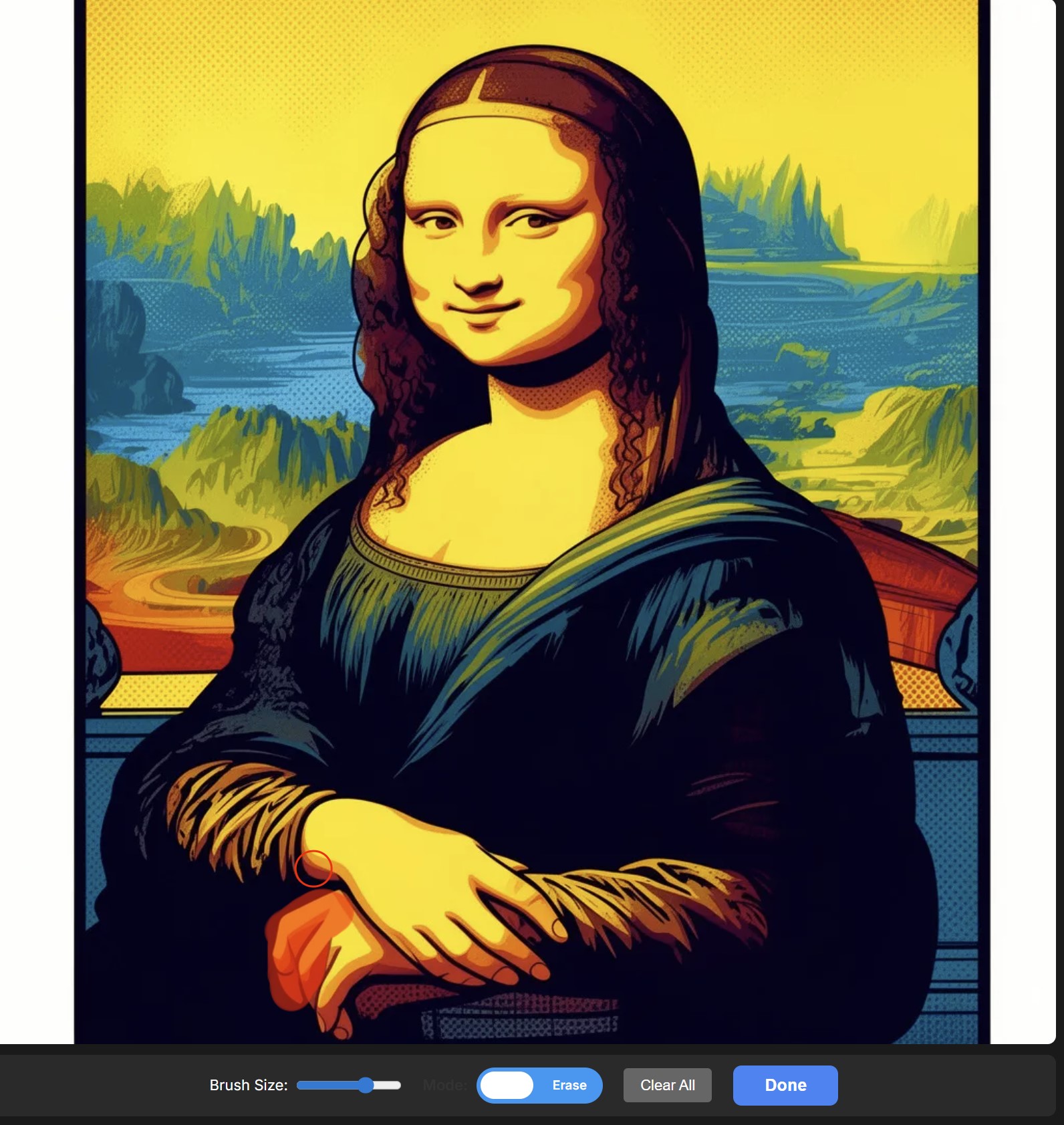}
    \caption{Artifact annotation tool. Annotators can brush over image regions that contain visual artifacts, enabling spatially localized artifact scoring.}
    \label{fig:eval_app_artifacts}
\end{figure}


\section{Annotation Examples}

\begin{figure}[!t]
    \centering
    \includegraphics[width=\linewidth]{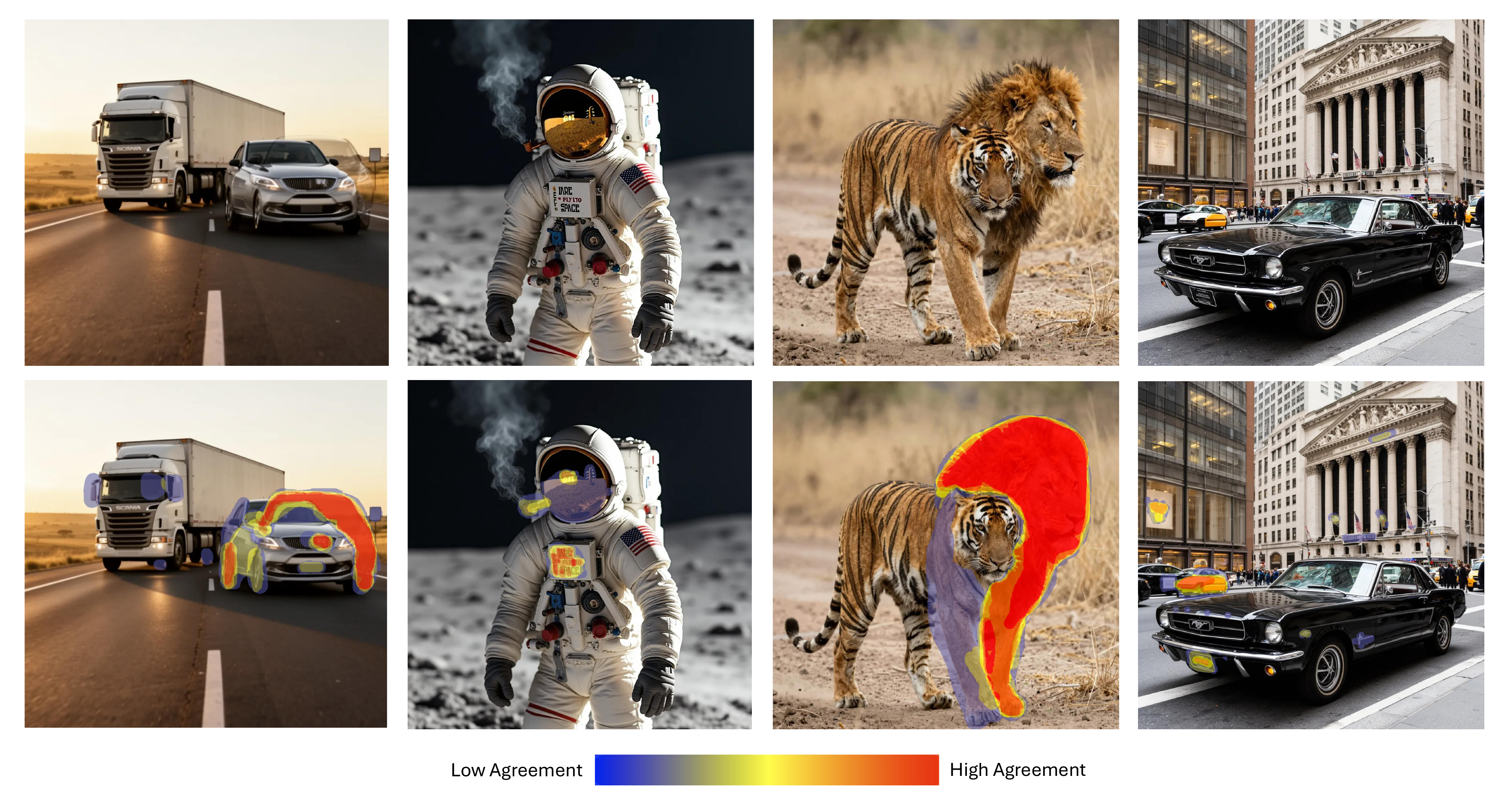}
    \caption{Examples of brush-based artifact annotations with inter-annotator agreement visualized as heatmaps. Red indicates strong agreement, while yellow and blue correspond to progressively weaker agreement.}
    \label{fig:artifact_examples}
\end{figure}

\myparagraph{Artifacts Brush}
\Cref{fig:artifact_examples} illustrates examples of brush-based artifact annotations together with the resulting inter-annotator agreement heatmaps. 
Regions with strong agreement (red) correspond to clear visual defects where annotators consistently identify artifacts. 
Areas with moderate agreement (yellow) typically reflect differences in how extensively annotators mark the spread of an artifact, which depends on how thoroughly the brush is applied. 
Finally, regions with low agreement (blue) indicate more subjective cases where some annotators label phenomena such as compression artifacts or reflections while others do not.
These ambiguities could be reduced with more detailed annotation guidelines. 
However, we intentionally kept the instructions minimal to mimic realistic crowd-sourced evaluation settings commonly used in text-to-image benchmarking.

\begin{figure}[!t]
    \centering
    \includegraphics[width=\linewidth]{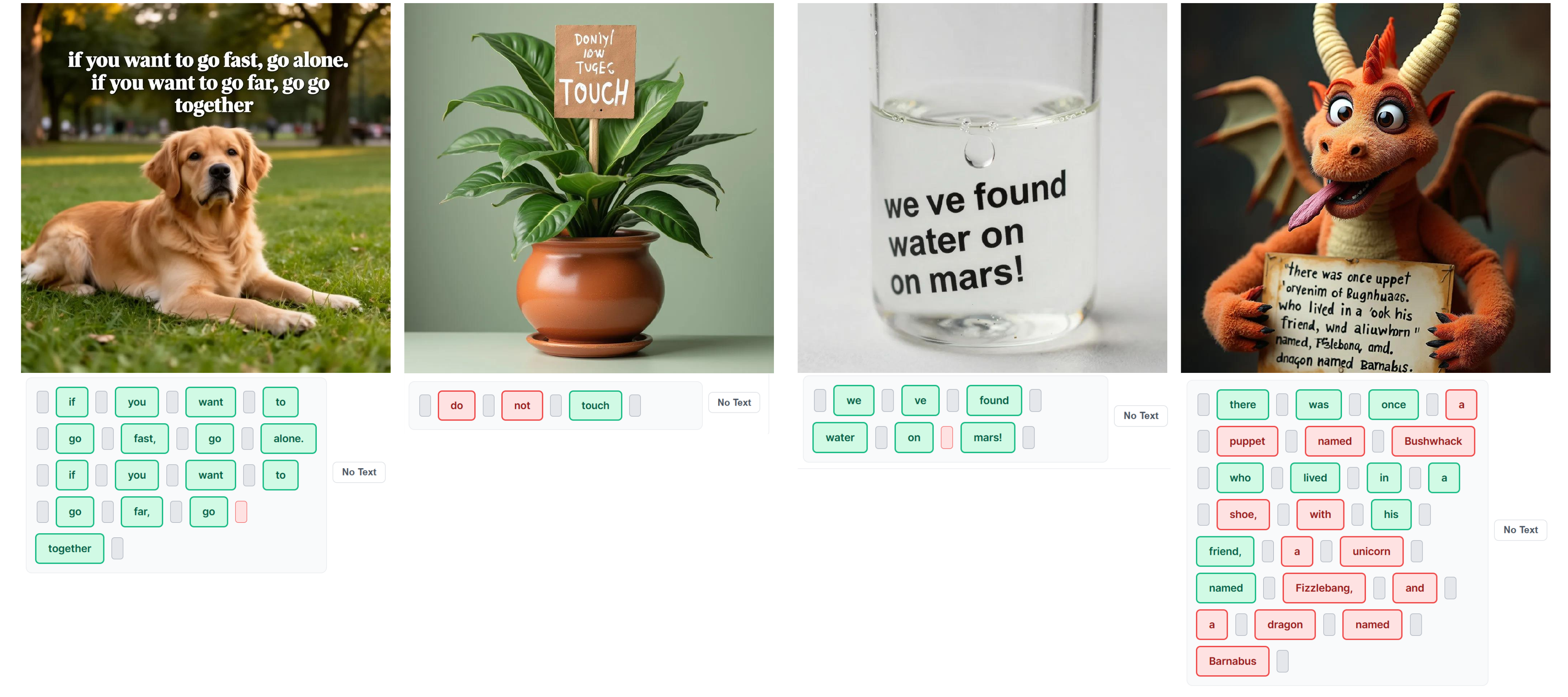}
    \caption{Examples of word-level text annotations illustrating correct words, missing words, and spurious insertions captured through gap annotations.}
    \label{fig:text_examples}
\end{figure}

\myparagraph{Word-Level Text}
\Cref{fig:text_examples} presents examples of word-level annotations for evaluating text rendering accuracy. 
Unlike standard word-level annotation, which cannot capture extra words inserted between requested tokens, our scheme explicitly includes \emph{gaps} between words. 
This allows annotators to identify spurious text that commonly appears in T2I outputs.
The word-level formulation also provides localized feedback by identifying which specific words fail to render correctly. 
This is particularly useful in challenging cases where correct text overlaps with gibberish, such as the dragon example in \Cref{fig:text_examples}. 
In such cases, automated approaches often fail to decode the text reliably, whereas human annotators can still accurately verify individual words.

\begin{figure}[!t]
    \centering
    \includegraphics[width=\linewidth]{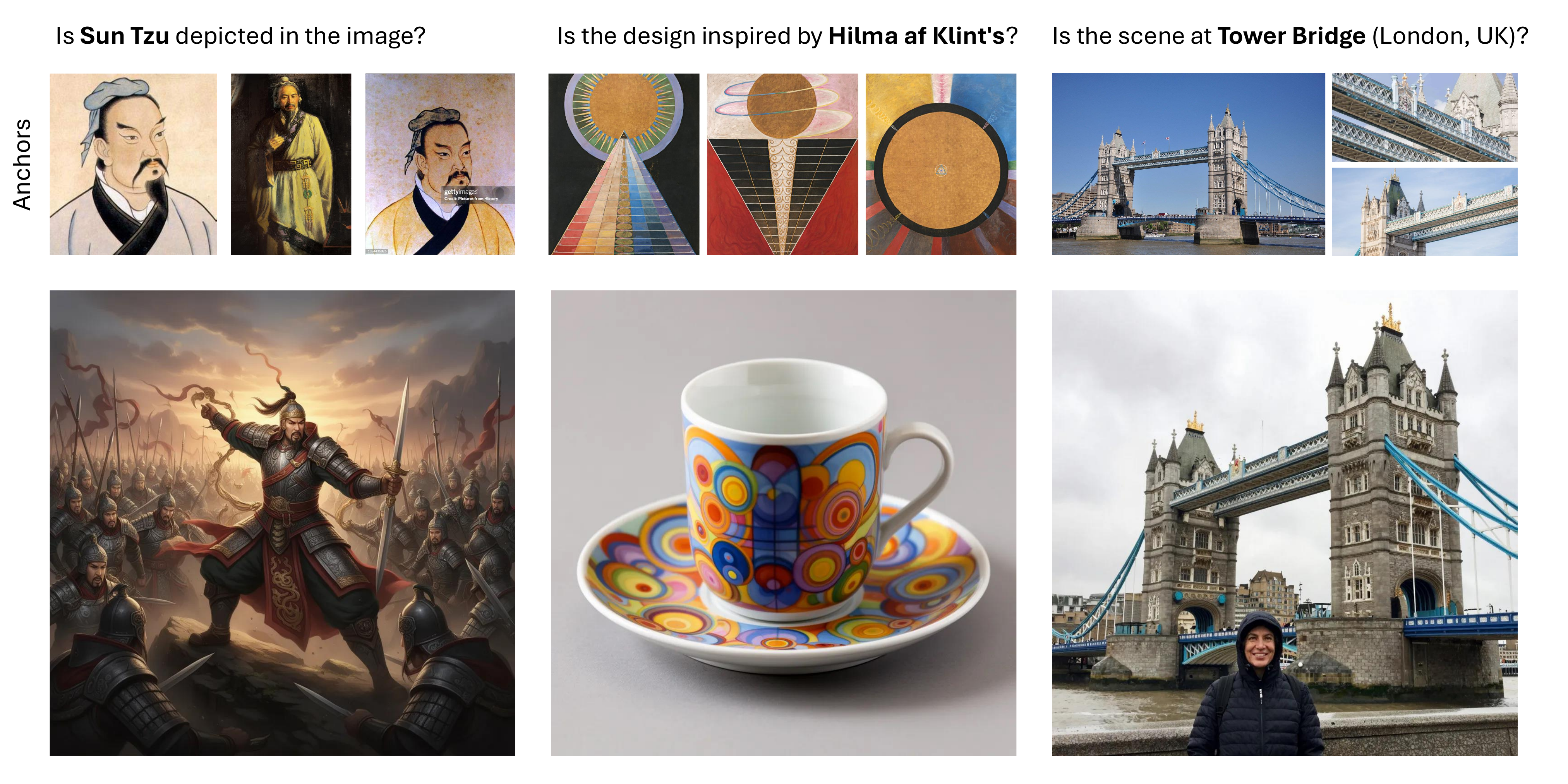}
    \caption{Examples of anchor references used to guide BQA or Likert scoring across categories such as characters, artwork styles, and landmarks.}
    \label{fig:anchor_examples}
\end{figure}

\myparagraph{Anchor Likert}
\Cref{fig:anchor_examples} presents examples of anchor images used for different skills. 
Directly answering some evaluation questions can be difficult for annotators who are not familiar with the subject. 
Providing reference anchors converts the task into a visual comparison problem, making it easier to judge the similarity between the generated image and known reference examples.

\section{Automated Evaluation Pipeline Details}

The automated evaluation pipeline mirrors the human protocol using a set of specialized agents, each implementing the annotation strategy associated with a particular skill. 
In total, the system includes four agents: Binary QA, Anchor Likert, Aesthetics Scoring, and Artifact Detection.

\myparagraph{(1) Binary QA Agent}
This agent evaluates factual skills using binary question answering (BQA), such as object presence, counting, and spatial relations. 
For each image, a batch of validation questions is provided together with unique identifiers (UIDs). 
The agent must respond to each question with \texttt{yes}, \texttt{no}, or \texttt{unsure}. 
To avoid ambiguity in object counting, the prompt explicitly instructs the model that if a prompt specifies \emph{a/an} object but the image contains multiple instances, the answer should be considered incorrect. 
The system prompt is shown below.

\begin{lstlisting}[style=jsonstyle]
# Answer the following questions about the image with only 'yes', 'no', or 'unsure'.
# For questions about presence of objects, if the prompt says a/an <object>,
# but the image has more, this should be flagged as wrong.
# UID: {question_uid_1}  Question: {question_text_1}
# UID: {question_uid_2}  Question: {question_text_2}
# ...
\end{lstlisting}
The output is programmatically validated to ensure that each UID is present and that the responses contain only valid labels.

\myparagraph{(2) Anchor Likert Agent}
This agent evaluates reference-dependent skills using the Anchor Likert protocol. 
The model receives a visual grid containing the generated image and two reference anchors that represent canonical examples of the evaluated concept. 
The agent is instructed to rate the similarity of the generated image to the reference examples on a scale from 0 to 5, following the same rubric used for human annotators.

\begin{lstlisting}[style=jsonstyle]
# You are evaluating a generated image against reference images.
# The leftmost image is the generated image. The other image(s) are reference examples.
# Question UID: {uid}
# Question: {question}
# Phrase being evaluated: "{phrase}"

# Rate how well the generated image matches the reference
# on a scale from 0 to 5:
# 0 = completely wrong / not present
# 1 = barely recognizable
# 2 = somewhat resembles
# 3 = moderately accurate
# 4 = mostly accurate
# 5 = perfectly matches
# Or answer 'unsure' if you cannot determine.
\end{lstlisting}

\myparagraph{(3) Aesthetics Score Agent}
This agent assesses the overall visual appeal of the generated image. 
The model is asked to rate the image on a 1--5 Likert scale reflecting perceived aesthetic quality, visual coherence, and attractiveness.

\begin{lstlisting}[style=jsonstyle]
# Rate the overall aesthetics, visual quality, and attractiveness of this image on a scale from 1 to 5:
# 1 = Very poor quality, unattractive, or heavily distorted
# 2 = Poor quality, noticeable flaws or unattractive
# 3 = Average quality, acceptable but not particularly impressive
# 4 = Good quality, visually pleasing with minor to no flaws
# 5 = Excellent quality, highly attractive and visually striking
\end{lstlisting}

\myparagraph{(4) Artifact Detection Agent}
Spatial artifact annotation cannot be directly produced by LLMs. 
Instead, we rely on pretrained artifact detection models to approximate the brush-based human annotations. 
In particular, we employ PAL4VST~\cite{zhang2023perceptual}, which produces pixel-level artifact masks that are used to compute artifact scores via mask area ratios. 
We also experimented with LEGION~\cite{kang2025legion}. 
However, we observed that it often highlights entire objects containing artifacts rather than localizing the artifact regions themselves. 
This behavior is incompatible with our definition of artifacts, which focuses on localized visual defects, and therefore PAL4VST was used in the final pipeline.




\section{Inter-LLM Agreement}
To further examine the consistency of the protocol across models, we measure inter-LLM agreement using Krippendorff's $\alpha$. 
The results are summarized in \Cref{tab:inter_llm_agreement}. 
Agreement is high for most structured skills, suggesting that different LLMs tend to produce consistent annotations under the proposed evaluation framework. 
In contrast, camera-related attributes, lighting, and aesthetic quality show lower agreement, reinforcing the observation that these skills remain challenging for automatic evaluation.


\begin{table}[!t]
\footnotesize
\centering
\begin{tabular}{lccccccccc}
\toprule
 & \faCube & \faSlidersH & \faBalanceScale & \faRunning & \faProjectDiagram & \faMountain & \faPaintBrush & \faLightbulb & \faCloudSun \\
\midrule
& 0.5884 
& 0.5830 
& 0.5397 
& 0.6662 
& 0.6989 
& 0.5743 
& 0.8040 
& 0.3158 
& 1.0000 \\
\midrule
 & \faEye & \faFont & \faSmile & \faIdBadge & \faLanguage & \faClock & \faCamera & \faStar & Avg.\\
\midrule
& 0.6455
& 0.8442
& 1.0000
& 0.7018
& 1.0000
& 0.4112
& 0.1026
& 0.2399
& 0.6303\\
\bottomrule
\end{tabular}
\vspace{5pt}
\caption{Inter-LLM Agreement (Krippendorff's Alpha) per Skill. LLMs exhibit moderate to high agreement across most skills except for lighting, camera, and aesthetic quality.
Icons correspond to: 
\faCube\ Entities, 
\faSlidersH\ Attributes, 
\faRunning\ Action, 
\faProjectDiagram\ Spatial Arrangement, 
\faBalanceScale\ Relative Comparison, 
\faLightbulb\ Lighting, 
\faCloudSun\ Weather, 
\faEye\ Viewpoint, 
\faCamera\ Camera Control, 
\faSmile\ Mood/Feeling, 
\faLanguage\ Language Complexity, 
\faClock\ Time, 
\faMountain\ Environment/Landmark, 
\faPaintBrush\ Style, 
\faIdBadge\ Named Entities, 
\faFont\ Text Rendering, 
and \faStar\ Aesthetic Quality.}
\label{tab:inter_llm_agreement}
\end{table}


\section{Challenging Scenarios}

\begin{figure}
    \centering
    \includegraphics[width=\linewidth]{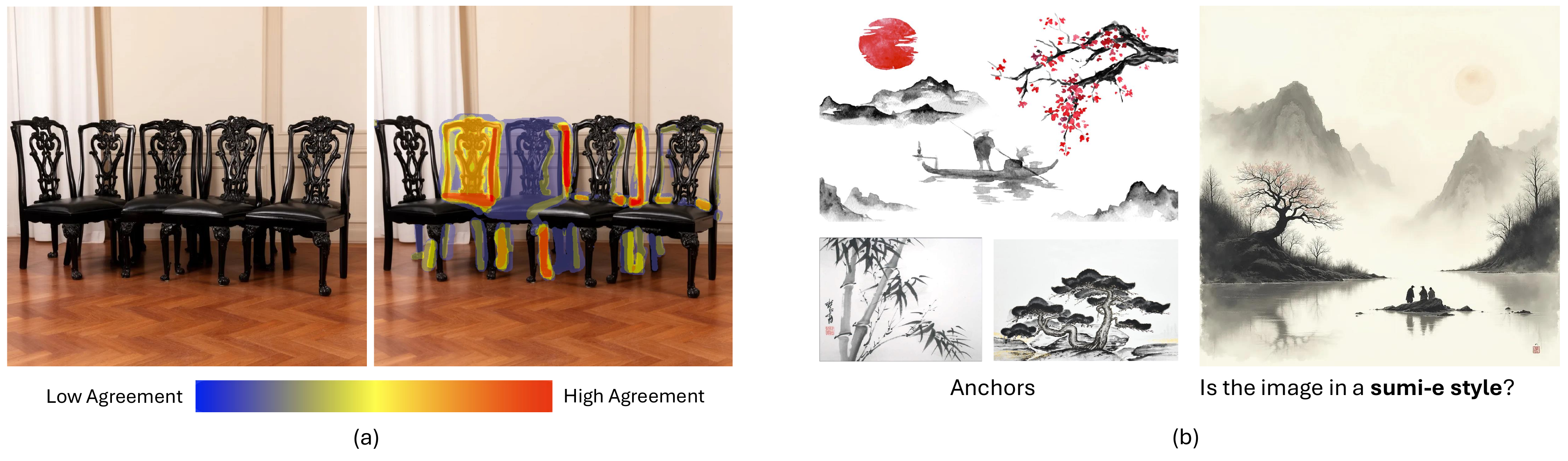}
    \caption{Examples of challenging annotation scenarios. (a) Overlapping chair legs create ambiguity in artifact localization. (b) Style matching remains difficult for non-expert annotators even with anchor references.}
    \label{fig:challenging}
\end{figure}

During our analysis, we observed several challenging cases for human annotation, illustrated in \Cref{fig:challenging}. 
In the chair example (a), the legs overlap heavily, making it difficult for annotators to determine which regions should be marked as artifacts. Despite this ambiguity, the agreement heatmap still reveals regions of strong consensus where the visual defect is clearly identifiable.

In the style example (b), even with anchor references, annotators without an art background may struggle to determine whether the generated image matches the target style. 
This difficulty arises because recognizing stylistic similarity often requires attention to specific attributes such as brush strokes, color intensity, or composition.
These challenges could be mitigated with clearer annotation guidelines or lightweight domain hints that help annotators focus on the relevant visual cues. 
Exploring such strategies is an interesting direction for improving annotation reliability in future work.


